\documentclass[10pt,twocolumn,letterpaper]{article}

\usepackage{wacv}              
\usepackage{comment}
\usepackage[accsupp]{axessibility}  
\usepackage{algorithm}
\usepackage{algpseudocode}
\usepackage{xcolor}
\definecolor{wacvblue}{rgb}{0.21,0.49,0.74}
\usepackage[pagebackref,breaklinks,colorlinks,allcolors=wacvblue]{hyperref}

\def\wacvPaperID{1013} 
\def\confName{WACV}
\def\confYear{2027}
\usepackage{ multirow, pifont,ctable,refcount}

\title{IDPrior: Decoupling Identity Generation from Facial Image Synthesis}

\author {
    Fadi Boutros\textsuperscript{\textrm 1},
    Meriç Demirörs\textsuperscript{\textrm 1 },
    Darian Tomašević\textsuperscript{\textrm 2},
    Vitomir Štruc\textsuperscript{\textrm 2},
     Naser Damer\textsuperscript{\textrm 1, \textrm 3},
\\
    \textsuperscript{\rm 1}Fraunhofer Institute for Computer Graphics Research IGD, Germany\\
     \textsuperscript{\rm 2}University of Ljubljana, Slovenia ,\textsuperscript{\rm 3}TU Darmstadt, Germany
    \\
    {\tt\small fadi.boutros@igd.fraunhofer.de}
}

\begin{document}
\maketitle
\begin{abstract}
Large-scale identity-labeled face datasets are essential for training modern face recognition models, yet collecting and annotating such data is costly and raises privacy and ethical concerns. Existing synthetic face generation methods typically derive identities from predefined class labels, unconditional image generation, or predefined latent sampling strategies, without explicitly modeling the underlying distribution of facial identities. In this paper, we propose IDPrior, a two-stage framework that decouples identity generation from image synthesis. We first learn a diffusion prior in the embedding space of a pretrained face recognition model to model the distribution of facial identities and generate novel identity embeddings. These embeddings are then used to condition a pretrained identity-conditioned diffusion model (DM), enabling the synthesis of multiple images of the same identity. This modular design provides a scalable approach to identity generation while remaining compatible with existing and future identity-conditioned DMs. Experiments indicate that IDPrior generates diverse and well-separated identities while maintaining a closer correspondence to the distribution of authentic identities. Face recognition models trained on datasets generated using IDPrior-generated identities achieve competitive or improved performance compared with existing synthetic-data-based methods across multiple face verification benchmarks.  Code and generated datasets are publicly available at \url{https://github.com/fdbtrs/IDperturb}.
\end{abstract}    
\section{Introduction}
\label{sec:intro}

Face recognition (FR) has witnessed remarkable progress over the past decade, driven by advances in deep neural networks, discriminative learning objectives, and large-scale identity-labeled datasets~\cite{Deng_2022,DBLP:conf/cvpr/Kim0L22,wang2018cosfacelargemargincosine}. However, collecting and annotating such datasets remains technically challenging and increasingly constrained by privacy regulations and ethical considerations~\cite{gdpr,lirias3838501,SyntheticFRSurvay}. The retraction of widely used datasets, including MS-Celeb-1M~\cite{guo2016ms} and VGGFace2~\cite{DBLP:conf/fgr/CaoSXPZ18}, further highlights the difficulty of acquiring authentic training data.

These challenges have motivated extensive research on synthetic data generation for FR~\cite{DBLP:journals/inffus/MelziTVKRLDMFOZZYZWLTKZDBVGFFMUG24,DBLP:conf/fgr/Otroshi-Shahreza24,DEANDRESTAME2025103099}. Recent GANs~\cite{Karras2020StyleGANADA} and diffusion models (DMs)~\cite{Rombach2021} can synthesize realistic faces for FR training. Existing approaches typically generate identities using class-conditional generators with discrete labels~\cite{Boutros2022SFace,10454585,DBLP:conf/cvpr/KolfREBKD23}, unconditional image generation~\cite{DBLP:conf/cvpr/Kim00023,DBLP:conf/iccv/BoutrosGKD23,DBLP:conf/iclr/LinHXMZD25}, or embedding-space identity generation based on predefined sampling strategies such as PCA~\cite{arc2face,DBLP:conf/iclr/WuST0B25} and \textcolor{black}{ virtual identity construction~\cite{vigface}.} However, these approaches do not explicitly learn a generative model of the underlying distribution of facial identities.

In this work, we propose IDPrior, a two-stage framework that decouples identity generation from image synthesis. We first learn a diffusion prior directly in the embedding space of a pretrained FR model to capture the distribution of facial identities and sample novel identity embeddings. These embeddings are then used as conditioning inputs for a pretrained identity-conditioned DM to synthesize multiple photorealistic images of each identity. This separation enables scalable identity generation while allowing the learned prior to be reused with existing and future identity-conditioned face generators without modifying their synthesis pipelines.

\noindent\textbf{Our contributions are:}
\begin{itemize}
\item We introduce IDPrior, a two-stage framework that decouples identity generation from image synthesis and can be seamlessly integrated with pretrained identity-conditioned face generators.
\item We propose a diffusion-based generative prior that explicitly learns the distribution of facial identities in the embedding space of a pretrained FR model.
\item We demonstrate that IDPrior provides a favorable trade-off between identity novelty, uniqueness, and distribution fidelity, enabling scalable generation of up to one million identities and improving downstream FR performance across multiple identity-conditioned generators.
\end{itemize}
\section{Related Work}
\label{sec:related_work}

Recent advances in GANs~\cite{Karras2020StyleGANADA} and DMs~\cite{Rombach2021} have enabled highly realistic face synthesis and stimulated extensive research on synthetic data for FR~\cite{DEANDRESTAME2025103099,DBLP:conf/fgr/Otroshi-Shahreza24}. Existing approaches can be broadly categorized by how they generate identities: class-label-based generation, unconditional image generation followed by identity extraction, and latent-space identity sampling.

\textbf{Class-label-based identity generation.}
The earliest synthetic face generation methods formulate identity generation as a class-conditional image synthesis problem. SFace~\cite{Boutros2022SFace}, SFace2~\cite{10454585}, and IDNet~\cite{DBLP:conf/cvpr/KolfREBKD23} adopt StyleGAN2-ADA~\cite{Karras2020StyleGANADA} to synthesize images conditioned on discrete identity labels. Since identities are represented by training class indices, these methods do not explicitly model the facial identity distribution and are limited by the available class labels (e.g., approximately 10K identities in SFace~\cite{Boutros2022SFace}). ExFaceGAN~\cite{ExFaceGAN} explores identity directions in the GAN latent space, but does not explicitly model the distribution of FR embeddings.

\textbf{Unconditional image generation followed by identity extraction.}
Recent diffusion-based methods generate faces unconditionally and extract identity embeddings using a pretrained FR network. DCFace~\cite{DBLP:conf/cvpr/Kim00023}, IDiff-Face~\cite{DBLP:conf/iccv/BoutrosGKD23}, ID$^3$~\cite{DBLP:conf/nips/Xu0WXDJHM0DH24}, and UIFace~\cite{DBLP:conf/iclr/LinHXMZD25} follow this paradigm, with additional mechanisms for identity consistency or appearance diversity. However, identities are implicitly determined by the image generator, limiting explicit control over identity distribution, separability, and scalability.

\textcolor{black}{
\textbf{Latent-space identity sampling.}
Another line of research generates or samples synthetic identities directly in
the FR embedding space before image synthesis.
Arc2Face~\cite{arc2face} trains Principal Component Analysis (PCA) on the
embedding space of a pretrained FR model and samples novel identity embeddings
from the resulting PCA space, which are then used to condition a pretrained
identity-conditioned diffusion model. Vec2Face~\cite{DBLP:conf/iclr/WuST0B25}
adopts the same PCA-based identity sampling strategy while introducing a
feature-masked autoencoder for image generation. IDiff-Face~\cite{DBLP:conf/iccv/BoutrosGKD23}
also explores direct sampling of identity embeddings from a standard Gaussian
distribution without explicitly modeling the empirical identity distribution.}

\textcolor{black}{
VIGFace~\cite{vigface} further explores identity generation in the FR embedding
space by explicitly constructing virtual identity prototypes and subsequently
using them to condition a diffusion-based face generator. In contrast to
sampling from a fixed statistical approximation, VIGFace optimizes virtual
prototypes to achieve separation from real identities while maintaining
identity consistency. This demonstrates the benefit of explicitly constructing
identities in the FR embedding space, but the resulting identity pool consists
of a predefined set of learned prototypes.}

While these methods establish the importance of generating identities in the FR embedding space, they might be suboptimal for learning identity distribution. PCA represents the identity distribution through a linear subspace and Gaussian sampling assumes a simple isotropic distribution, which may not capture the complex nonlinear structure of facial identity embeddings. VIGFace instead learns individual virtual identity prototypes, resulting in a predefined prototype pool, and expanding the pool requires additional prototype training.
IDPrior takes a different approach by learning a generative model directly over the empirical identity-embedding distribution and sampling identities from this learned prior. This formulation enables IDPrior to generate new identities without being restricted to a predefined prototype pool while capturing nonlinear distributional structure beyond a fixed PCA approximation.

\begin{figure*}
    \centering
    \includegraphics[width=0.85\linewidth]{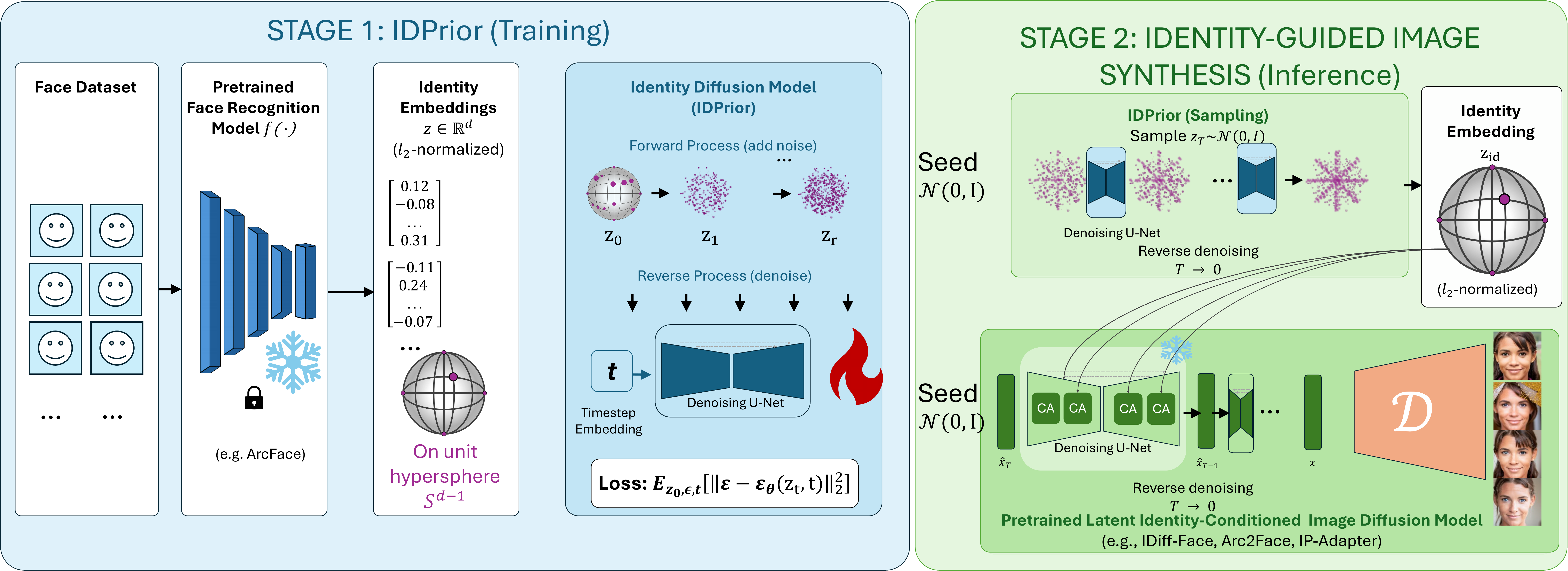}
    \vspace{-2mm}
   \caption{
Overview of IDPrior, a dual diffusion framework that decouples identity generation from image synthesis. Stage 1: A diffusion prior is learned on FR embeddings to model the distribution of facial identities and generate novel identity embeddings. Stage 2: The sampled embeddings are subsequently used to condition a pretrained identity-guided image diffusion model, enabling the synthesis of multiple images per identity. 
}
    \label{fig:overview}
    \vspace{-5mm}
\end{figure*}
\section{Methodology}
\label{sec:methodology}

We propose a two-stage framework for synthesizing novel face identities and
their corresponding images. We first learn the distribution of facial identity
embeddings in the embedding space of a pretrained FR model. Specifically, we
train a denoising diffusion probabilistic model (DDPM)~\cite{Ho2020} to model
the distribution of normalized identity embeddings extracted from real face
images. Once trained, the model is used to sample novel identity embeddings.
These embeddings are subsequently used to condition a pretrained
identity-guided image DM~\cite{DBLP:conf/iccv/BoutrosGKD23,ye2023ip-adapter,IDperturb,arc2face},
which synthesizes multiple face images belonging to the same identity. By
decoupling identity generation from image synthesis, the proposed framework
enables scalable generation of large identity-consistent face datasets
(Fig.~\ref{fig:overview}).

\subsection{Problem Formulation}

Let
\(
x\in\mathcal{X}\subset\mathbb{R}^{H\times W\times3}
\)
denote a face image sampled from the real image distribution
\(p_{\mathrm{data}}(x)\).
A pretrained FR model
\(f:\mathcal{X}\rightarrow\mathbb{R}^{d}\)
maps each image to a \(d\)-dimensional identity embedding
\(z=f(x)\). Following modern FR models
\cite{Deng_2022,wang2018cosfacelargemargincosine,DBLP:conf/cvpr/Kim0L22},
all embeddings are $\ell_2$ normalized, $\|z\|_2=1$, such that the embedding
space lies on the unit hypersphere \(\mathcal{S}^{d-1}\).

Given a training set of face images, we extract the corresponding embeddings:
\begin{equation}
\mathcal{Z}
=
\{z_i=f(x_i)\}_{i=1}^{N},
\end{equation}
which induces the empirical distribution of FR embeddings
\(p_{\mathrm{data}}(z)\). Our objective is to learn a generative model that
samples novel embeddings while preserving the statistical properties of this
distribution.

We model the synthetic-data generation process as the following two-stage factorization:
\begin{equation}
p(x,z)
=
p_{\theta}(z)\,
p_{\phi}(x|z),
\end{equation}
where \(p_{\theta}(z)\) models the distribution of facial identity embeddings
and \(p_{\phi}(x|z)\) generates face images conditioned on a sampled identity.

FR embeddings provide a compact representation in which identity similarity is
encoded through angular distances, making them a natural domain for modeling
facial identity independently of image appearance. IDPrior therefore learns
the identity distribution directly in the embedding space, separating identity
generation from image synthesis and enabling scalable sampling of novel
identity embeddings.

\subsection{Identity Diffusion Model}

To model the distribution of identity embeddings, we employ a DDPM~\cite{Ho2020}.
Let \(z_0\) denote a clean identity embedding sampled from
\(p_{\mathrm{data}}(z)\). The forward diffusion process gradually perturbs the
embedding by adding Gaussian noise:
\begin{equation}
q(z_t|z_{t-1})
=
\mathcal{N}
\left(
\sqrt{1-\beta_t}\,z_{t-1},
\beta_t I
\right),
\end{equation}
where \(\{\beta_t\}_{t=1}^{T}\) is a predefined variance schedule.
Equivalently,
\begin{equation}
z_t
=
\sqrt{\bar{\alpha}_t}\,z_0
+
\sqrt{1-\bar{\alpha}_t}\,\epsilon,
\end{equation}
where
\[
\epsilon\sim\mathcal{N}(0,I),
\qquad
\alpha_t=1-\beta_t,
\qquad
\bar{\alpha}_t
=
\prod_{s=1}^{t}\alpha_s.
\]

Although the clean identity embeddings lie on the unit hypersphere, diffusion is performed in the ambient Euclidean space following the standard DDPM formulation. The final generated embeddings are subsequently normalized to ensure compatibility with the FR embedding space.

The reverse diffusion process is parameterized by a neural network
\(\epsilon_{\theta}(z_t,t)\) that predicts the injected Gaussian noise. The
reverse transition is defined as
\begin{equation}
p_{\theta}(z_{t-1}|z_t)
=
\mathcal{N}
\left(
z_{t-1};
\mu_{\theta}(z_t,t),
\sigma_t^2I
\right),
\end{equation}
where
\begin{equation}
\mu_{\theta}(z_t,t)
=
\frac{1}{\sqrt{\alpha_t}}
\left(
z_t
-
\frac{1-\alpha_t}
{\sqrt{1-\bar{\alpha}_t}}
\epsilon_{\theta}(z_t,t)
\right).
\end{equation}

\textbf{Training Objective:}
The identity diffusion model is trained using the standard noise-prediction
objective~\cite{Ho2020}:
\begin{equation}
\mathcal{L}_{\mathrm{id}}
=
\mathbb{E}_{z_0,\epsilon,t}
\left[
\|
\epsilon-
\epsilon_{\theta}(z_t,t)
\|_2^2
\right],
\end{equation}
where \(t\) is sampled uniformly from \(\{1,\ldots,T\}\).

\subsection{Sampling Novel Identity Embeddings}
After training, novel identities are generated by sampling
$z_T\sim\mathcal{N}(0,I)$ and iteratively applying the learned reverse
diffusion process:
\begin{equation}
z_{t-1}
\sim
p_{\theta}(z_{t-1}|z_t),
\qquad
t=T,\ldots,1.
\end{equation}

The final denoised embedding $\hat{z}_0$ is $\ell_2$-normalized:
\begin{equation}
z_{\mathrm{id}}
=
\frac{\hat{z}_0}
{\|\hat{z}_0\|_2},
\end{equation}
ensuring compatibility with the normalized embedding space of the pretrained
FR model. The resulting embedding represents a novel synthetic identity and
is subsequently used as a condition for image generation.

\subsection{Identity-Conditioned Image Generation}

Given an identity embedding \(z_{\mathrm{id}}\), we synthesize face images
using a pretrained identity-conditioned DM (Stage 2 in
Fig.~\ref{fig:overview}). The image generator models the conditional
distribution \(p_{\phi}(x|z_{\mathrm{id}})\), where the identity embedding is
provided as the conditioning signal. Keeping \(z_{\mathrm{id}}\) fixed while
varying the random seed produces multiple images with the same target identity
and different appearances.

\begin{figure}
    \centering
    \includegraphics[width=0.80\linewidth]{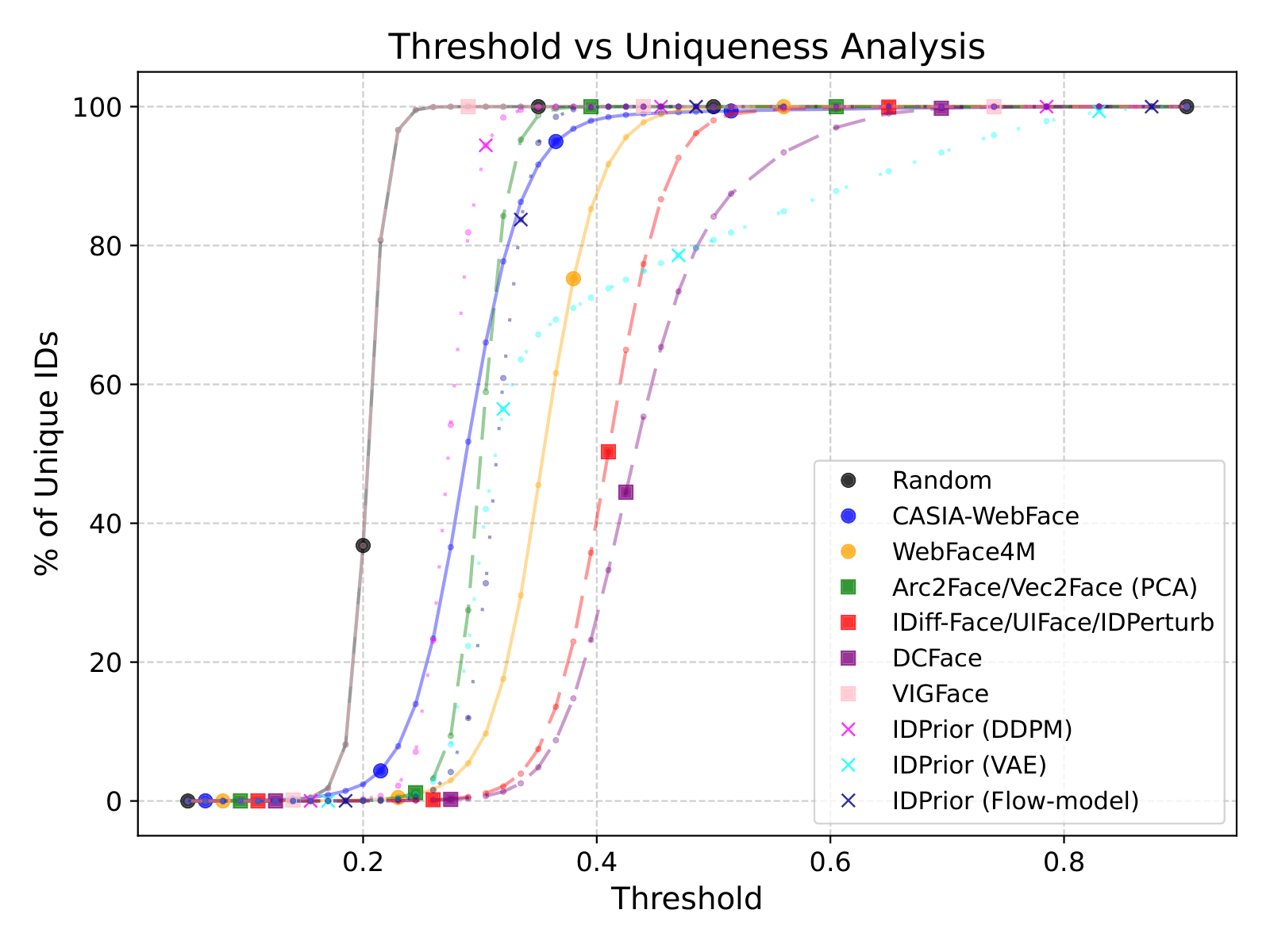}
    \vspace{-3mm}
    \caption{
    \color{black}Identity uniqueness as a function of the cosine-similarity threshold. We report the percentage of unique identities among 10K generated identities. 
    As the threshold becomes stricter, IDPrior (DDPM) consistently preserves a larger fraction of unique identities than existing approaches.
    }
    \label{fig:Identity_uniqueness}
    \vspace{-3mm}
\end{figure}

\section{Experimental Setup}
\vspace{-2mm}
\subsection{IDPrior Implementation Details}
\textbf{IDPrior Training:} IDPrior is trained on normalized face recognition embeddings extracted from the WF4M~\cite{DBLP:conf/cvpr/ZhuHDY0CZYLD021} dataset (4M images). We train separate IDPrior models using a pretrained ArcFace~\cite{Deng2020} backbone for integration with Arc2Face and IP-Adapter-FaceID, and a pretrained ElasticFace~\cite{ElasticFace} backbone for integration with IDiff-Face and \textsc{IDperturb}. 
\textcolor{black}{Since IDPrior models the identity distribution in the embedding space of a pretrained FR model, its performance and generated identity distribution inherently depend on the underlying FR representation.}

\textcolor{black}{ 
To assess whether the effectiveness of IDPrior depends on the diffusion
formulation, we additionally instantiate the identity prior using a
flow-based model \cite{DBLP:conf/iclr/LipmanCBNL23} and a variational autoencoder (VAE) \cite{DBLP:journals/corr/KingmaW13}. All variants are trained on the same normalized
identity embeddings and used to sample identities for image generation. 
Unless otherwise specified, the DDPM is the main IDPrior model. 
}

Each identity embedding has 512 dimensions and is reshaped into a $16\times32$ feature map to facilitate the use of a commonly utilized two-dimensional U-Net architecture~\cite{Rombach2021}. We adopt the DDPM~\cite{Ho2020} with a linear noise schedule over $T=1000$ diffusion timesteps~\cite {Ho2020,Rombach2021}. During training, a timestep is uniformly sampled, Gaussian noise is added according to the forward diffusion process, and the network is optimized to predict the injected noise, as described in Section~\ref{sec:methodology}.
The U-Net~\cite{Rombach2021} consists of five encoder and five decoder stages with channel dimensions ${64,128,256,512,512}$. Each stage contains two residual blocks, the attention head dimension is set to $8$, and the network uses a dropout rate of $0.1$ throughout. The model contains 114M trainable parameters, \textcolor{black}{ 342 GFLOPs (171 GMACs)} and is trained on four NVIDIA A100 GPUs using mixed-precision training, requiring approximately 14 hours. 
Training is performed for 100 epochs using the Adam optimizer~\cite{AdamOptimizer} with a learning rate of $1\times10^{-4}$, a batch size of $512$, and a cosine annealing learning-rate schedule~\cite{DBLP:conf/iclr/LoshchilovH17}. During inference, we generate novel identity embeddings using a DDIM sampler~\cite{DBLP:conf/iclr/SongME21} with 50 denoising steps, enabling the synthesis of one million identities, each  of dimension 512, in approximately two hours (i.e., 7.2 ms per sample). 
\textcolor{black}{These costs are incurred offline, and the identity prior is not required during subsequent image synthesis. The generated embeddings can be stored and reused across different pretrained generators, amortizing the upfront cost for large-scale synthetic data generation.}

\textbf{Image Synthesis:}
We evaluate IDPrior with IDiff-Face~\cite{DBLP:conf/iccv/BoutrosGKD23},
\textsc{IDperturb}~\cite{IDperturb}, Arc2Face~\cite{arc2face}, and
IP-Adapter-FaceID~\cite{ye2023ip-adapter}, replacing their original identity
generation mechanisms with IDPrior-sampled embeddings. IP-Adapter-FaceID is
a dual-conditional (image/text) DM not originally designed for synthetic FR
training. We additionally evaluate a PCA baseline trained on ArcFace
embeddings from WF4M. 


\subsection{Face recognition training and evaluation}


\textbf{FR Training Setup:}
We train ResNet50~\cite{He2015DeepRL} with CosFace~\cite{wang2018cosfacelargemargincosine}
on the synthetic datasets, following~\cite{DBLP:conf/iccv/BoutrosGKD23,ExFaceGAN,DBLP:conf/iclr/LinHXMZD25,DBLP:conf/nips/SunSPT24}.
We use SGD optimizer for 25 epochs with batch size 512, momentum 0.9, weight decay
$5\times10^{-4}$, initial learning rate 0.1, and cosine annealing learning rate scheduler.
The CosFace \cite{wang2018cosfacelargemargincosine} margin and scale are 0.35 and 64, respectively.

\textbf{Evaluation Benchmarks:}
FR performance is reported, following  \cite{DBLP:conf/iccv/BoutrosGKD23,ExFaceGAN,DBLP:conf/iclr/LinHXMZD25,DBLP:conf/cvpr/Kim0L22}, as the verification accuracy on: Labeled Faces in the Wild LFW \cite{LFWDatabase}, AgeDB-30 \cite{AgeDB30Database}, Cross-Age LFW (CA-LFW) \cite{CALFWDatabase}, Celebrities in Frontal-Profile in the Wild (CFP-FP) \cite{CFPFPDatabase}, and Cross-Pose LFW (CP-LFW) \cite{CPLFWDatabase}, following their official evaluation protocol.

\begin{figure}[t!]
    \centering
    \begin{subfigure}[b]{0.85\linewidth}
        \centering
        \includegraphics[width=\linewidth]{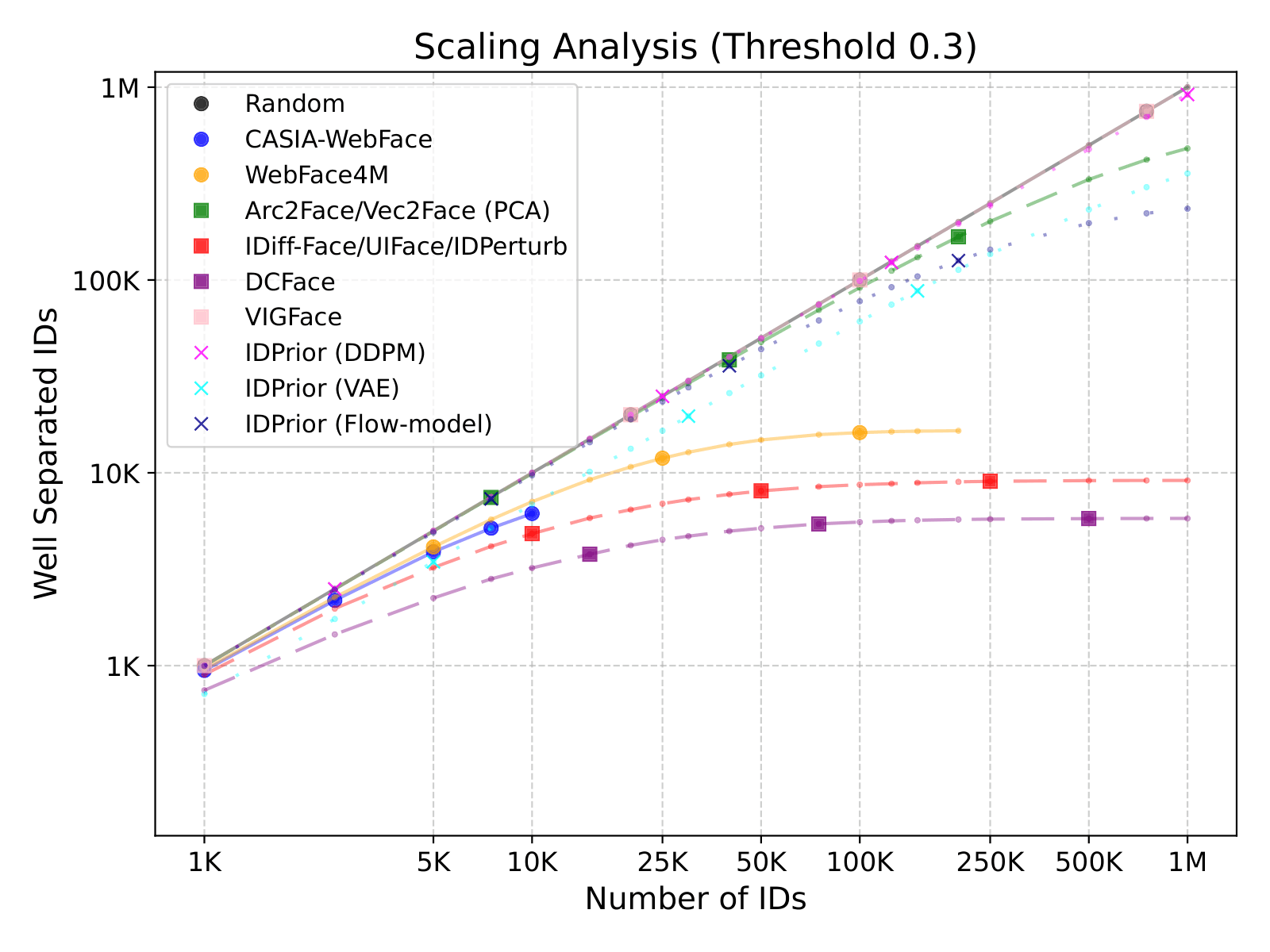}
                                \vspace{-7mm}
        \caption{}
    \end{subfigure}
    \vspace{-3mm}
    \caption{Number of well-separated identities as the total number of generated identities increases, using a threshold of  0.30. 
    }
\label{fig:identity_scalable}
\vspace{-5mm}
\end{figure}

\section{Results}
\label{sec:result}
\vspace{-1mm}
\subsection{Unique Identity Generation}
To evaluate the identity uniqueness, we compare it with synthetic identity sampling methods, including DCFace~\cite{DBLP:conf/cvpr/Kim00023}, IDiff-Face~\cite{DBLP:conf/iccv/BoutrosGKD23}, \textsc{IDperturb}~\cite{IDperturb}, UIFace~\cite{DBLP:conf/iclr/LinHXMZD25}, Arc2Face~\cite{arc2face}, and Vec2Face~\cite{DBLP:conf/iclr/WuST0B25}. DCFace, IDiff-Face, \textsc{IDperturb}, and UIFace implicitly generate identities by synthesizing face images and treating each image as a new identity, whereas Arc2Face and Vec2Face explicitly sample identity embeddings from a PCA model learned on FR embeddings. We also evaluate normalized Gaussian vectors, which are used by IDiff-Face~\cite{DBLP:conf/iccv/BoutrosGKD23} as reference identity embeddings.
Given a set of generated identity embeddings $\{z_i\}_{i=1}^{N}$, we define the set of unique identities as:
\begin{equation}
U=
\left\{
z_i:\;
sim(z_i,z_j)<r,\;
\forall j<i
\right\},
\end{equation}
where $sim(\cdot,\cdot)$ denotes the cosine similarity and $r$ is a similarity threshold. We vary $r\in[0,1]$ to evaluate identity uniqueness under different matching criteria. As references, we additionally report the uniqueness statistics of the authentic CW~\cite{DBLP:journals/corr/YiLLL14a} and WF4M~\cite{DBLP:conf/cvpr/ZhuHDY0CZYLD021}.
Figure~\ref{fig:Identity_uniqueness} shows that IDPrior (DDPM) achieves
substantially higher identity uniqueness than existing image-generation-based
and PCA-based methods across the evaluated similarity thresholds. While
random sampling and VIGFace can achieve near-perfect
separation, high uniqueness alone does not imply a realistic identity
distribution. As shown in the next subsection, these methods exhibit
substantially worse distribution fidelity than IDPrior (DDPM).

\begin{table}[t!]
\centering
\caption{
Scalability of identity generation from 10K to 1M identities. We report the percentage of unique identities at thresholds 0.25, 0.30, and 0.40. 
IDPrior (DDPM) maintains substantially higher uniqueness than PCA- and image-generation-based approaches across scales, while random sampling, VIGFace, and the flow-based prior
achieve higher separation at the cost of poorer distribution fidelity, as detailed in Table \ref{tab:identity_unique}.
}
\vspace{-3mm}
\resizebox{0.83\linewidth}{!}{%
\begin{tabular}{lllll}
\multirow{2}{*}{Approach}                    & \multirow{2}{*}{\#Identities} & \multicolumn{3}{l}{Unique in (\%) @$r$ } \\
                                             &                               & 0.25     & 0.3      & 0.4     \\ \hline

\multirow{4}{*}{Random (uniform)}                           & 10K                           & 99.98   & 100.00    & 100.00\\
                                             & 100K                          & 99.98   & 100.00 & 100.00  \\
                                             & 500K                          & 99.88  & 100.00 & 100.00  \\
                                             & 1M                            & 99.77 & 100.00 & 100.00   \\ \hline \hline

\multirow{4}{*}{\color{black} VIGFace \cite{vigface}}                     & \color{black} 10K                           & \color{black}100.00   & \color{black}100.00    &\color{black}100.00 \\
                                             & \color{black} 100K                          & \color{black} 99.98  &\color{black} 100.00   & \color{black}100.00\\
                                             & \color{black} 500K                          &\color{black} 99.88  &\color{black} 99.99 &\color{black}100.00  \\
                                             & \color{black} 1M                            &\color{black} 99.76   &\color{black}99.99   & \color{black}100.00 \\ \hline                                             
\multirow{1}{*}{Arc2Face \cite{arc2face}}                           & 10K                           & 74.35    & 99.15    & 100.00\\
\multirow{1}{*}{Vec2Face \cite{DBLP:conf/iclr/WuST0B25} }                                              & 100K                          & 16.55   & 91.25   & 99.99  \\
                                              & 500K                          & 3.34   & 66.48  & 99.99  \\
                                             & 1M                            & 1.67   & 48.15  & 99.99  \\ \hline
\multirow{4}{*}{ DCFace \cite{DBLP:conf/cvpr/Kim00023} }                      & 10K                           & 8.11     & 32.10     & 83.75   \\
                                             & 100K                          & 0.83     & 5.55    & 61.42  \\
                                             & 500K                          & 0.17    & 1.16   & 36.48 \\
                                             & 1M                            & 0.08    & 0.58     & 26.34  \\ \hline
\multirow{1}{*}{IDiff-Face \cite{DBLP:conf/iccv/BoutrosGKD23} } & 10K                           & 11.79    & 48.35    & 97.6    \\
\multirow{1}{*}{UIFace \cite{DBLP:conf/iclr/LinHXMZD25}}                                             & 100K                          & 1.23    & 8.68     & 82.43  \\
  \multirow{1}{*}{\textsc{IDperturb} \cite{IDperturb}}                                           & 500K                          & 0.25   & 1.82    & 54.38 \\
                                             & 1M                            & 0.12   & 0.91   & 40.43 \\ \hline

\multirow{4}{*}{\color{black} IDPrior (VAE)}                     & \color{black} 10K                            &\color{black} 49.87    &\color{black} 68.58    & \color{black} 77.15\\
                                             & \color{black} 100K                           & \color{black}13.80   &\color{black} 60.97   &\color{black} 75.00\\
                                             & \color{black} 500K                         & \color{black}2.85  &\color{black} 46.38  &\color{black} \color{black}73.56 \\
                                             & \color{black} 1M                           & \color{black}1.42   &\color{black} 35.72  &\color{black} \color{black}72.95 \\ 
                                             
                                             \hline
\multirow{4}{*}{\color{black} IDPrior (Flow-model)}                     & \color{black} 10K                           & \color{black}54.05   & \color{black}97.04     &\color{black} 99.99 \\
                                             & \color{black} 100K                          &\color{black} 8.30   & \color{black}77.60   &\color{black} 99.99 \\
                                             & \color{black} 500K                          & \color{black}1.66  & \color{black}39.46 & \color{black}99.97 \\
                                             & \color{black} 1M                            &\color{black} 0.83   &\color{black} 23.44  &\color{black} 99.95 \\

                                             \hline
\multirow{4}{*}{IDPrior (DDPM)}                     & 10K                           & 93.44    & 99.91    & 100.00\\
                                             & 100K                          & 59.39   & 98.89   & 100.00\\
                                             & 500K                          & 20.40  & 95.39  & 99.99 \\
                                             & 1M                            & 10.68   & 91.63  & 99.99 \\

                                             \hline                                             
\end{tabular}}

\label{tab:Id_uniqueness_details}
\end{table}

\subsection{Scalability of Identity Generation}
To evaluate the scalability, we analyze how the number of well-separated identities evolves as more identities are generated. Given a set of generated identity embeddings ${z_i}_{i=1}^{N}$, an embedding $z_i$ is considered well-separated if:
\begin{equation}
\max_{j \neq i} sim(z_i, z_j) < r.
\end{equation}
We progressively increase the number of generated identities up to one million and count the number of embeddings satisfying this criterion.

Figure~\ref{fig:identity_scalable} compares IDPrior with random Gaussian embeddings, DCFace~\cite{DBLP:conf/cvpr/Kim00023}, IDiff-Face~\cite{DBLP:conf/iccv/BoutrosGKD23}, \textsc{IDperturb}~\cite{IDperturb}, UIFace~\cite{DBLP:conf/iclr/LinHXMZD25}, Arc2Face~\cite{arc2face}, VIGFace \cite{vigface}, Vec2Face~\cite{DBLP:conf/iclr/WuST0B25}, and the authentic CW~\cite{DBLP:journals/corr/YiLLL14a} and WF4M~\cite{DBLP:conf/cvpr/ZhuHDY0CZYLD021} datasets using thresholds $r=0.3$. The threshold $r=0.3$ follows~\cite{DBLP:conf/cvpr/Kim00023,IDperturb} and corresponds to the operating point of the evaluation model~\cite{Deng_2022}, which reports a verification accuracy of 97.17\% at a threshold of 0.3080 using $TPR@FPR = 0.01\%$ on the IJB-B benchmark \cite{inproceedingsijbb}. $r=0.25$ provides a more stringent criterion.

At one million generated identities and $r=0.3$, IDPrior retains 91.63\% well-separated identities, compared to less than 1\% for image-generation-based methods and 48.15\% for PCA-based approaches, as shown in  Table~\ref{tab:Id_uniqueness_details}. 
Although random embeddings yield the highest separability due to their near-uniform hyperspherical distribution, they do not reflect real facial identity distributions, as indicated by their poor ID-FID (Section \ref{sec:id_FID}). Overall, these results show that explicitly learning the facial identity distribution provides substantially better scalability than image-generation- or PCA-based identity sampling.
The trends observed in Figure~\ref{fig:identity_scalable} are further quantified in Table~\ref{tab:Id_uniqueness_details}. 

\begin{table*}[t!]
\centering
\caption{
\textcolor{black}{Identity uniqueness, distribution fidelity, and potential identity leakage on 10K generated identities. We report  uniqueness at  thresholds of 0.30 and 0.25, ID-FID, unified ID-FID (U-ID-FID) using a common CelebA \cite{celeba} reference distribution, and nearest-training-identity similarity statistics, including the $95$th and $99$th percentiles ($P_{95}$ and $P_{99}$) and leakage rates at $\tau$ of 0.25, 0.3, and 0.4. Lower $P_{95}$, $P_{99}$, and leakage rates indicate lower similarity to training identities and therefore greater identity novelty, while lower ID-FID and U-ID-FID indicate better distributional fidelity.}}

\vspace{-3mm}
\resizebox{\linewidth}{!}{%
\begin{tabular}{lllllllllll}
\hline
                              & \multicolumn{2}{c}{Unique}                         & \multicolumn{2}{c}{ID-FID}                  & \multicolumn{1}{c}{Novelty} & \multicolumn{5}{c}{Identity leakage $\downarrow$ }              \\ \hline
Method                        & Unique$@$0.3 $\uparrow$ & Unique$@$0.25 $\uparrow$ & ID-FID $\downarrow$ & \color{black} U-ID-FID $\downarrow$ & Train Sim. $\downarrow$     & \color{black}$P_{95}$    &\color{black} $P_{99}$    &\color{black} Leak$@$0.25 (\%)  &\color{black} Leak$@$0.3 (\%) &\color{black} Leak$@$0.4 (\%)  \\ \hline
CW (Authentic)     & 61.50                   & 16.67                    & -                   & \color{black} 0.0309                     & -                           & -      & -      & -         & -        & -        \\
WF4M (Authentic)         & 70.83                   & 22.68                    & -                   &\color{black} 0.0560                     & -                           & -      & -      & -         & -        & -        \\ \hline
Random (Uniform)              & 100.00                  & 99.98                    & 0.6009              & \color{black}0.3210                & 0.1666                      &\color{black} 0.1932 & \color{black}0.2075 & \color{black}0.00      & \color{black}0.00     & \color{black}0.00     \\

\color{black} VIGFace & \color{black} 100.00 & \color{black}100.00  &\color{black}0.9166 &\color{black}0.9987 &\color{black} 0.1605 &\color{black} 0.1903 &\color{black} 0.2052 &\color{black} 0.00 &\color{black} 0.00  &\color{black} 0.00 \\
Arc2Face /Vec2Face            & 99.15                   & 74.35                    & 0.6759              & \color{black}0.6967                & 0.1686                      & \color{black}0.1936 &\color{black} 0.2087 &\color{black} 0.01      &\color{black} 0.00     &\color{black} 0.00     \\
IDiff-Face /UIFace /\textsc{IDperturb} & 48.35                   & 11.79                    & 0.0271              & \color{black}0.0421                & 0.3159                      & \color{black}0.3810 & \color{black}0.4240 &\color{black} 98.64     & \color{black}64.45    & \color{black}2.47     \\
DCFace                        & 32.10                   & 8.11                    & 0.0267              & \color{black}0.0769                & 0.3497                      & \color{black}0.4872 & \color{black}0.5604 &\color{black} 99.07     &\color{black} 78.72   & \color{black}17.70   \\

\color{black} IDPrior (VAE)                & \color{black} 68.58                  & \color{black} 49.87                   & \color{black} 0.2728            & \color{black}\color{black} 0.2256              & \color{black} 0.2201                    & \color{black} 0.2715 &\color{black} 0.2957 &\color{black} 17.40   &\color{black} 0.74   &\color{black} 0.00    \\ 

\color{black} IDPrior  (Flow-model)               & \color{black} 97.04                  & \color{black} 54.05                 & \color{black} 0.1088            & \color{black}\color{black} 0.0631              & \color{black} 0.2482                    & \color{black} 0.2885 &\color{black} 0.3122 &\color{black} 42.07   &\color{black} 2.21   &\color{black}  0.00   \\ 

IDPrior (DDPM)                & 99.91                   & 93.44                    & 0.0261              & \color{black}0.0557                & 0.2213                      & \color{black}0.2534 &\color{black} 0.2715 &\color{black} 6.64     &\color{black} 0.05   &\color{black} 0.00     \\ 
\hline

\hline
\end{tabular}
}
\vspace{-4mm}
\label{tab:identity_unique}
\end{table*}

\subsection{Identity Distribution Similarity}
\label{sec:id_FID}
To evaluate how well the generated identities match the distribution of real identities, we compute the Identity Fréchet Distance (ID-FID) between real and synthetic embeddings. Unlike the conventional FID, which measures image realism, ID-FID directly evaluates the similarity between identity distributions.
Given the real and synthetic embedding sets $\mathcal{Z}_{r}$ and $\mathcal{Z}_{syn}$ with Gaussian statistics $(\mu_r,\Sigma_r)$ and $(\mu_{syn},\Sigma_{syn})$, ID-FID is defined as:
\begin{equation}
\mathrm{ID\mbox{-}FID}
=
\|\mu_r-\mu_{syn}\|_2^2
+
\operatorname{Tr}
\left(
\Sigma_r+\Sigma_{syn}
-
2(\Sigma_r\Sigma_{syn})^{1/2}
\right).
\end{equation}

\textcolor{black}{
The ID-FID values in Table~\ref{tab:identity_unique} are computed against the authentic training distribution of each method. To remove this dependency on the reference dataset, we additionally report a unified ID-FID (U-ID-FID) using CelebA \cite{celeba} as a common reference for all
methods. IDPrior (DDPM) achieves the lowest ID-FID and second-lowest U-ID-FID of 0.0261 and 0.0557, respectively, while retaining 99.91\% uniqueness at $r=0.30$, confirming that its distributional fidelity is not solely due to evaluation against its own training distribution. In contrast, VIGFace and random sampling achieve near-perfect separation, but substantially higher U-ID-FID, illustrating that identity separation alone does not imply a realistic identity distribution.
We further compare alternative generative priors under the same
identity-embedding setting. The flow-based prior achieves 97.04\% uniqueness and U-ID-FID of 0.1088, whereas the VAE performs worse in both uniqueness and distribution fidelity.
Overall, the IDPrior (DDPM) provides the best combination of separation and
 fidelity among the evaluated priors.
}

\subsection{Novelty with Respect to Training Identities}
We evaluate the novelty of generated identities with respect to the training dataset. To this end, we measure their similarity to the identities in the training set.

Let
$\mathcal{Z}_{\mathrm{train}}=\{z_i^{\mathrm{train}}\}_{i=1}^{N}$
denote the identity embeddings extracted from the training dataset, and
$\mathcal{Z}_{\mathrm{gen}}=\{z_i^{\mathrm{gen}}\}_{i=1}^{M}$
denote the generated identity embeddings. For each generated identity, we compute its maximum cosine similarity to all training identities,
\begin{equation}
s_i
=
\max_{j}\{
sim\left(
{z_i^{\mathrm{gen}}},
z_j^{\mathrm{train}}
\right)\}.
\end{equation}

Lower similarity indicates greater identity novelty and, correspondingly, a lower potential for identity leakage from the training set. Table~\ref{tab:identity_unique} reveals a trade-off between novelty, uniqueness, and distribution fidelity. PCA, random sampling, and VIGFace exhibit lower training
similarity, but their substantially higher ID-FID and U-ID-FID indicate that this
novelty partly results from deviating from the authentic identity distribution. Conversely, image-generation-based methods better match the distribution but exhibit substantially higher training similarity
and severe identity collisions. IDPrior (DDPM) provides a favorable
balance, with 99.91\% uniqueness at $r=0.30$ and a U-ID-FID of 0.0557.


\begin{figure}
    \centering
    \includegraphics[width=0.80\linewidth]{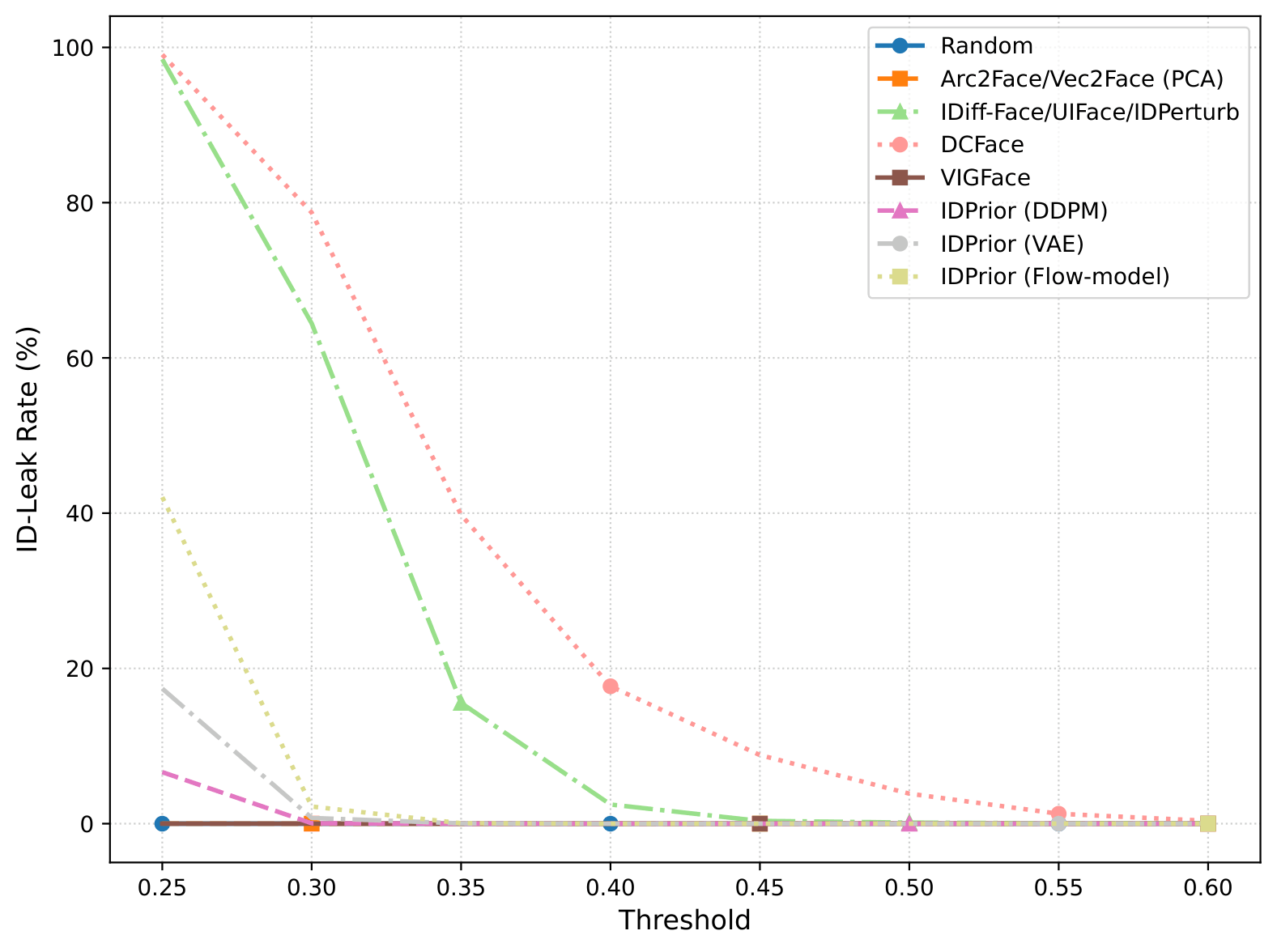}
    \vspace{-3mm}
    \caption{
    \color{black}Identity leakage at different  thresholds. We report the percentage of leaked identities among 10K generated identities. 
    }
    \label{fig:Identity_leak}
    \vspace{-4mm}
\end{figure}

\subsubsection{Identity leakage analysis}
To quantify potential identity leakage, we report the leakage rate at a similarity threshold $\tau$. For each generated identity $z_i^{\mathrm{gen}}$, we compute its maximum cosine similarity to the training identities and count it as a potential leakage case if this similarity exceeds $\tau$ of 0.25, 0.3, or 0.4. The leakage rate is defined as:
\begin{equation}
\mathrm{Leakage@}\tau =
\frac{1}{M}
\sum_{i=1}^{M}
\mathbb{I}
\left[
\max_{z_j^{\mathrm{train}} \in \mathcal{Z}_{\mathrm{train}}}
\cos(z_i^{\mathrm{gen}},z_j^{\mathrm{train}}) > \tau
\right],
\end{equation}
where $\mathbb{I}[\cdot]$ is the indicator function. Lower leakage rates indicate fewer generated identities that are highly similar to individual training identities, suggesting lower potential for identity leakage. We additionally report the 95th and 99th percentile (noted as $P_{95}$ and $P_{99}$) of the nearest neighbor (NN) between each item in $\mathcal{Z}_{\mathrm{gen}}$ and  $\mathcal{Z}_{\mathrm{train}}$. Results for thresholds 0.25 and 0.3 are presented in Table \ref{tab:identity_unique} and for the full range [0.25,0.6] quantified in Figure \ref{fig:Identity_leak}. 
For IDPrior (DDPM), only 0.05\% of generated identities exceed a cosine similarity of 0.30 with any training identity, and none exceed 0.40.
Thus, although distribution-aware sampling yields higher similarity than purely random, PCA,  or virtual prototyping sampling, high-similarity cases indicative of potential identity leakage remain rare.


\begin{table}[t!]
\caption{
\textcolor{black}{
Comparison of identity generation strategies for training FR models. Verification accuracy (\%) obtained using different identity sources with identity-conditioned DMs. 
Replacing the original identity sampling mechanisms with IDPrior generally improves downstream recognition performance across the evaluated identity-conditioned generators, with the largest gains observed for IDiff-Face and IP-Adapter-FaceID.
Results marked with * are reproduced in this work.}
}
\vspace{-2mm}
\centering
\resizebox{\linewidth}{!}{%
\begin{tabular}{llllllll}
\hline
Face Embeddings            & DGM        & LFW   & AgeDB-30 & CFP-FP & CA-LFW & CP-LFW & Avg   \\ \hline
Uncond. Samples & IDiff-Face \cite{DBLP:conf/iccv/BoutrosGKD23} & 98.75 & 88.85    & 91.61  & 90.90  & 86.15  & 91.25 \\

Random (Uniform) & IDiff-Face \cite{DBLP:conf/iccv/BoutrosGKD23} & 98.00 & 87.30 & 90.97 & 91.25 & 86.01 & 90.71\\
VIGFace & IDiff-Face \cite{DBLP:conf/iccv/BoutrosGKD23} & 99.82 &87.48 & 91.39 &90.07 & 86.53 & 91.06 \\
\color{black}IDPrior(VAE)               & \color{black}IDiff-Face \cite{DBLP:conf/iccv/BoutrosGKD23} & \color{black}98.60 & \color{black}88.72    & \color{black}91.53  & \color{black}91.45  &\color{black} 87.27 & \color{black}91.51 \\ 
\color{black}IDPrior (Flow-model)               & \color{black}IDiff-Face \cite{DBLP:conf/iccv/BoutrosGKD23} &\color{black} 99.03 & \color{black}90.95    & \color{black} 92.30  &\color{black} 91.90  &\color{black} 87.03 &\color{black} 92.24 \\ 

IDPrior (DDPM)               & IDiff-Face \cite{DBLP:conf/iccv/BoutrosGKD23} & 98.82 & 90.80    & 92.67  & 91.95  & 87.07  & 92.26 \\ \hline
Uncond. samples  & \textsc{IDperturb} \cite{IDperturb}  & 99.40 & 93.20    & 93.61  & 93.50  & 88.37  & 93.62 \\
IDPrior               & \textsc{IDperturb} \cite{IDperturb}  &  99.23      & 93.43         &    93.97    &   93.35     &   89.70     &93.94     \\ \hline
PCA*  & Arc2Face* \cite{arc2face}  & 96.58 & 84.46    & 88.62  & 86.58  & 78.85  & 87.02  \\
IDPrior               & Arc2Face \cite{arc2face}   &   97.70    &  80.57        &  89.17      &  84.75      &     83.77   &  87.19     \\ \hline
PCA*  & IP-Adapter-FaceID*   & 93.43 &    74.42 & 79.41   &  78.12 & 73.45 & 79.77  \\
IDPrior               & IP-Adapter-FaceID   &94.37      &    79.82     &    81.76  &  81.70     &     74.63   &  82.46     \\ \hline

\end{tabular}}
\label{tab:FR_generalizability}
\vspace{-3mm}
\end{table}

\subsection{Synthetic-Data-based Face Recognition}
\textbf{Integration with Identity-Conditioned DMs:}
To evaluate IDPrior's compatibility with different identity-conditioned generators, we replace the original identity sampling mechanisms of IDiff-Face~\cite{DBLP:conf/iccv/BoutrosGKD23}, \textsc{IDperturb}~\cite{IDperturb} and Arc2Face~\cite{arc2face} with identity embeddings sampled from IDPrior while keeping all other components unchanged. For IDiff-Face, we integrated the three priors, DDPM, VAE, and flow-based models.
We further integrate our IDPrior with IP-Adapter-FaceID~\cite {ye2023ip-adapter}, a dual conditional (text and image) DM model that is not originally designed for generating synthetic data for FR development.

\begin{table}[t!]
\centering
\caption{
FR performance trained on IDPrior (DDPM)-generated datasets of increasing scale.  Performance generally improves, approaching the accuracy achieved using authentic training data.
}
\vspace{-2mm}
\resizebox{\linewidth}{!}{%
\begin{tabular}{llllllll}
\hline
           &  IDs × imgs/ID    & LFW   & AgeDB-30 & CFP-FP & CA-LFW & CP-LFW & Avg \\ \hline
IDPrior & 50K (1k x 50)  & 94.83 & 79.82    & 79.07  & 84.20  & 73.80  & 82.34 \\
IDPrior &  0.5M (10k x 50) & 98.82 & 90.80    & 92.67  & 91.95  & 87.07  & 92.26 \\
IDPrior &  1.0M (20k x 50) & 99.27 & 92.28    & 94.03  & 92.98  & 88.90  & 93.49 \\
IDPrior &  2.0M (40k x 50) & 99.40 & 93.72    & 94.79  & 93.93  & 90.20  & 94.41 \\
IDPrior &  4.0M (80k x 50) & 99.52 & 94.83  & 95.83  & 94.48  &  90.98& 95.13 \\
\hline
\end{tabular}}
\label{tab:FR_scalablility}
\end{table}

Table~\ref{tab:FR_generalizability} shows that IDPrior generally improves downstream FR performance across all image generation models. 
In particular, replacing the identities sampled by IDiff-Face and Arc2Face with ones generated by IDPrior yields notable gains on most benchmarks. 
IDPrior (Flow-model) and IDPrior (DDPM) achieve very competitive FR performance, with DDPM obtaining the best results on CFP-FP and CA-LFW.
These results demonstrate that the quality of the generated identities plays a crucial role in the effectiveness of synthetic face datasets and validate the modular design of IDPrior as a plug-and-play identity generator for identity-conditioned diffusion models.

\begin{table*}[th!]
\centering
\caption{
FR verification accuracies of IDPrior and existing synthetic-data-based approaches. At 0.5M and 1M images, IDPrior achieves competitive or SOTA performance among the
compared methods. The first and second rows report authentic CW and WF4M as references. 
Under each setup, the best results are in \textbf{bold} and the second-best ones are \underline{underlined}. \textcolor{black}{Results in the last block are obtained by augmenting authentic CW\cite{DBLP:journals/corr/YiLLL14a} with synthetic data, following VIGFace~\cite{vigface}.  Results marked with  \dag  are obtained from \cite{vigface}}.}
\vspace{-2mm}
\resizebox{0.94\textwidth}{!}{%
\begin{tabular}{c|cccccccc}
\hline
  \multirow{2}{*}{\textbf{Method}} &
  \multirow{2}{*}{\textbf{\#imgs (IDs × imgs/ID)}} &
  \multirow{2}{*}{\textbf{LFW}} &
  \multirow{2}{*}{\textbf{AgeDB-30}} &
  \multirow{2}{*}{\textbf{CFP-FP}} &
  \multirow{2}{*}{\textbf{CA-LFW}} &
  \multirow{2}{*}{\textbf{CP-LFW}} &
  \multirow{2}{*}{\textbf{Avg}}  \\
  &&&&&&&\\ \hline
CW \cite{DBLP:journals/corr/YiLLL14a,DBLP:conf/iccv/BoutrosGKD23} &
  0.49M($\sim$10.5k IDs) & 
  99.55 &
  94.55 &
  95.31 &
  93.78 &
  89.95 &
  94.63  \\ 
  WF4M \cite{DBLP:conf/cvpr/ZhuHDY0CZYLD021} &
  4M($\sim$206K IDs) & 
  99.80 &
  97.80 &
  99.03 &
  96.03 &
  94.27 &
  97.39  \\ 
  \hline
DigiFace \cite{DigiFace1M} WACV'22   & 0.5M(10k × 50)   & 95.40 & 76.97 & 87.40 & 78.62 & 78.87 & 83.45 \\ 

IDnet \cite{DBLP:conf/cvpr/KolfREBKD23} CVPRW'23    & 0.53M(10.5k × 50)  & 92.58 & 73.53 & 75.40  & 79.90 & 74.25 & 79.13 \\
SFace \cite{Boutros2022SFace}  IJCB'22           & 0.63M(10.5k × 60) & 91.87 & 71.68 & 73.86 & 77.93 & 73.20  & 77.71 \\
SFace2+ \cite{10454585}   TBIOM'24                    & 0.63M(10.5k × 60) & 95.60  & 77.37 &  77.11  & 83.40 & 74.60 & 81.62 \\
SynFace \cite{DBLP:conf/iccv/QiuYG00T21} ICCV'21        &0.5M(10k × 50)   & 88.98 & -    & -     & -     & -     & -\\
SynFace (w/IM) \cite{DBLP:conf/iccv/QiuYG00T21} ICCV'21  &0.5M(10k × 50)   & 91.93 & 61.63 & 75.03 & 74.73 & 70.43 & 74.75    \\
USynthFace  \cite{DBLP:conf/fgr/BoutrosKFKD23} FG'23  & 0.4M(400k × 1)   & 92.23 & 71.62 & 78.56  & 77.05 & 72.03 & 78.30 \\
ExFaceGAN(Con)     \cite{ExFaceGAN}   IJCB'23       & 0.5M(10k × 50)   & 93.50  & 78.92  & 73.84 & 82.98  & 71.60 & 80.17 \\

Vec2Face \cite{DBLP:conf/iclr/WuST0B25} ICLR'25 & 0.5M(10k × 50) & 98.87 & 93.12 & 88.97 & 93.57& 85.47 & 92.00 \\

DCFace  \cite{DBLP:conf/cvpr/Kim00023}  CVPR'23  & 0.5M(10k × 50) & 98.55 & 89.70 & 85.33 & 91.60 & 82.62 & 89.56 \\ 
ID$^3$ \cite{DBLP:conf/nips/Xu0WXDJHM0DH24}  NeurIPS'24 & 0.5M(10k × 50) & 97.68 & 91.00 & 86.84 & 90.73 & 82.77 & 89.80 \\

CemiFace \cite{DBLP:conf/nips/SunSPT24}  NeurIPS'24& 0.5M(10k × 50) & 99.03 & 91.33 & 91.06 & 92.42& 87.65 & 92.30 \\

NegFaceDiff \cite{DBLP:journals/corr/abs-2508-09661} ICCV-W'25   & 0.5M(10k × 50)  & 98.98 &	90.02 &	91.67 &	91.65	& 88.82&	92.23 \\
\color{black}VIGFace \cite{vigface} ICCV'25 & \color{black}0.5M(10k × 50)   &\color{black} 99.02 &\color{black} 90.95 & \color{black}95.09 &\color{black}90.00 & \color{black}87.72 &\color{black} 92.56 \\
UIFace \cite{DBLP:conf/iclr/LinHXMZD25}  ICLR'25 & 0.5M(10k × 50) &99.27 & 90.95 & 94.29 & 92.25 & 89.58 &93.27 \\
IDiff-Face \cite{DBLP:conf/iccv/BoutrosGKD23,IDperturb} ICCV'23 & 0.5M(10k × 50)  &
98.75 & 88.85    & 91.61  & 90.90  & 86.15  & 91.25 
\\ 

HyperFace \cite{DBLP:conf/iclr/Otroshi-Shahreza25} ICLR'25  & 0.5M(10k × 50)   &
98.50 &  86.53	&  88.83	&89.40 & 84.23  &89.49 \\

\textsc{IDperturb} \cite{IDperturb} CVPR'26    & 0.5M(10k × 50)  &
99.40&  	93.20	&  93.61	& 93.50&  	88.37&  	\underline{93.62} \\ 

IDPrior - IDiff-Face (Ours)   & 0.5M(10k × 50) & 98.82 & 90.80    & 92.67  & 91.95  & 87.07  & 92.26\\ 
IDPrior - \textsc{IDperturb} (Ours)   & 0.5M(10k × 50)  &
  99.23      & 93.43         &    93.97    &   93.35     &   89.70     &\textbf{93.94} 
\\ \hline
DigiFace \cite{DigiFace1M} WACV'22  & 1.2M(10k × 72 + 100k × 5)   &
96.17 &  81.10	&  89.81	& 82.55 & 82.23 &  86.37 \\
DCFace \cite{DBLP:conf/cvpr/Kim00023}  CVPR'23   & 1.2M(20k × 50 + 40k × 5)   &
98.58 &  90.97	& 88.61 	&  92.82 & 85.07 &  91.21 \\

Arc2Face \cite{arc2face}  ECCV'24    & 1.2M(20k × 50 + 40k × 5)  &
98.92 &  92.45	& 94.58 	& 93.33 &  86.45 &  93.14 \\
CemiFace  \cite{DBLP:conf/nips/SunSPT24}  NeurIPS'24   & 1.0M(20k × 50)  &
99.18 &  91.97	& 92.75 	& 93.01 & 88.42 & 93.07 \\
\color{black} VIGFace \cite{vigface} ICCV'25 &\color{black} 1.2M(60k × 20) & \color{black}99.48 & \color{black}93.62&  \color{black}97.07 &\color{black} 92.88 &\color{black}90.15  &\color{black}\underline{94.64} \\
Vec2Face \cite{DBLP:conf/iclr/WuST0B25}  ICLR'25   & 1.0M(20k × 50 )   &
98.87 &  93.85	& 89.87 	& 93.65 & 86.13 &  92.47 \\

UIFace \cite{DBLP:conf/iclr/LinHXMZD25} ICLR'25  & 1.0M(20k × 50)   &
99.22 &  92.45	&  95.03	& 93.18 & 90.42 &94.06 \\

\textsc{IDperturb}  \cite{IDperturb} CVPR'26    & 1.0M(20k × 50)  &

99.48	 & 94.03 & 	95.01 & 	93.85 & 	90.01 & 94.48 \\

IDPrior - IDiff-Face (Ours)   & 1.0M(20k × 50)  
& 99.27 & 92.28    & 94.03  & 92.98  & 88.90  & 93.49 	 \\ 
IDPrior - \textsc{IDperturb} (Ours)   & 1.0M(20k × 50)  & 99.47
&  		94.57 &  95.07	& 94.00 &  90.15	& \textbf{94.65} 	 \\  
\hline
UIFace \cite{DBLP:conf/iclr/LinHXMZD25} ICLR'25  & 1.5M(30k × 50)   &
99.38 & 93.28	& 95.96 	& 93.43 & 90.67 &94.54 \\
\color{black} VIGFace \cite{vigface} ICCV'25 & \color{black} 6.0M(120k × 50) & \color{black} 99.33 &\color{black} 93.82 &\color{black} \color{black} 97.31 & \color{black} 92.95 & \color{black} 91.12 & \color{black} \underline{94.91} \\
Vec2Face \cite{DBLP:conf/iclr/WuST0B25}  ICLR'25   & 5.0M(20k × 50 )   &
99.25 & 94.38	&90.36	& 94.12 &86.75 &  92.97 \\

Vec2Face \cite{DBLP:conf/iclr/WuST0B25}  ICLR'25   & 20.0M(400k × 50 )   &
99.37 & 94.33	& 90.53 	& 94.60  & 87.35 &  93.24 \\

HyperFace \cite{DBLP:conf/iclr/Otroshi-Shahreza25} ICLR'25  & 3.2M(50k × 64)   &
98.27 &  90.40		&92.24& 91.48  & 85.60 & 91.59\\

IDPrior - IDiff-Face (Ours) & 4.0M (80K x 50)  & 99.52 & 94.83  & 95.83  & 94.48  &  90.98 & \textbf{95.13} \\
\hline \hline

\color{black} CW + IDiff-Face \cite{DBLP:conf/iccv/BoutrosGKD23} \cite{vigface}\dag 
& \color{black} 0.49M + 0.5M 
&\color{black}  99.58
&\color{black}  94.78
&\color{black}  97.04
&\color{black}  94.00
&\color{black}  90.40
&\color{black}  95.16 \\

\color{black} CW + GANDiffFace \cite{vigface}\dag 
& \color{black} 0.49M + 0.5M 
&\color{black}  99.52
&\color{black}  93.98
&\color{black}  96.61
&\color{black}  93.57
&\color{black}  90.30
&\color{black}  94.80 \\
\color{black} CW + Vec2Face \cite{DBLP:conf/iclr/WuST0B25} \cite{vigface}\dag 
&\color{black}  0.49M + 1M
&\color{black}  99.60
&\color{black}  95.48
&\color{black}  97.41
&\color{black}  94.40
&\color{black}  91.07
&\color{black}  95.59 \\

\color{black} CW + DigiFace \cite{DigiFace1M}   \cite{vigface}\dag     &   \color{black} 0.49M + 1.2M  &\color{black}  99.37
& \color{black} 94.95
& \color{black} 97.51
& \color{black} 93.77 &
\color{black} 90.92&
\color{black} 95.30 \\
\color{black} CW + DCFace \cite{DBLP:conf/cvpr/Kim00023} \cite{vigface}\dag & \color{black}   0.49M + 1.2M 

&\color{black}  99.43
& \color{black} 95.20
&\color{black}  96.97
&\color{black}  94.38
&\color{black}  90.33
&\color{black}  95.26 \\

\color{black} CW + VIGFace \cite{vigface} \dag 
&\color{black}  0.49M + 1M
& \color{black} 99.55
& \color{black} 95.73
& \color{black} 98.03
& \color{black} 94.12
&\color{black}  91.80
& \color{black} 95.85\\

\color{black} CW + VIGFace + Synthetic CW \cite{vigface} \dag 
&\color{black}  0.49M + 1M + 0.2M
&\color{black}  99.70
&\color{black}  95.85
&\color{black}  98.10
&\color{black}  94.38
&\color{black}  91.57
&\color{black}  \underline{95.92} \\

\color{black} CW + IDPrior 
&\color{black}  0.49M + 0.5M &\color{black}  99.45 & \color{black} 95.43 & \color{black} 97.50 &\color{black}  94.13 \color{black} &\color{black} 91.28 &\color{black}  95.56 \\
\color{black} CW + IDPrior &\color{black}  0.49M + 1M & \color{black} 99.67 & \color{black} 95.60 & \color{black} 97.73 & \color{black} 94.23 & \color{black} 91.83 &\color{black}  95.81 \\
\color{black} CW + IDPrior &\color{black}  0.49M + 4M &\color{black}  99.62 &\color{black}  96.45 &\color{black}  98.19 & \color{black} 95.08 &\color{black}  92.07 &\color{black}  \textbf{96.28} \\

\hline
\end{tabular}
}
\label{tab:sota}
\vspace{-2mm}
\end{table*}

\textbf{Scaling Synthetic FR Training:}
We progressively increase the number of synthetic identities generated by IDPrior and train FR models on datasets ranging from 1K to 80K identities.
The results, reported in Table~\ref{tab:FR_scalablility}, show a consistent improvement in recognition performance as the number of generated identities increases.
\textcolor{black}{
Increasing the dataset scale generally improves FR performance.}

\textbf{Comparison with SOTA Synthetic Face Datasets:}
We compare our IDPrior with state-of-the-art (SOTA) approaches on common FR benchmarks. The comparison includes both GAN-based and diffusion-based methods trained exclusively on synthetic data.
We integrate IDPrior into IDiff-Face \cite{DBLP:conf/iccv/BoutrosGKD23} and \textsc{IDperturb} \cite{IDperturb} by replacing their original identity generation strategy with identities sampled from the proposed diffusion prior. We evaluate both models on synthetic datasets with 0.5M and 1M identities to ensure fair, comparable results against SOTA.

Table \ref{tab:sota} shows that IDPrior achieves competitive or SOTA performance among the compared synthetic-data-based approaches, while using fewer synthetic images in several comparisons.
Replacing the original identity sampling mechanism of IDiff-Face and \textsc{IDperturb} with IDPrior generally improves recognition performance for both dataset sizes, 0.5M and 1M (common scale for synthetic-based FR \cite{IDperturb,DBLP:conf/iclr/LinHXMZD25}).
Few works reported the results of larger scale, and those results are included at the end of Table \ref{tab:sota}. 
This demonstrates that identity generation constitutes an independent bottleneck in synthetic face generation pipelines and highlights the effectiveness of decoupling identity generation from image synthesis.
\textcolor{black}{We further evaluate IDPrior as a data-augmentation strategy by combining synthetic identities with authentic CASIA-WebFace (CW), following \cite{vigface}. Adding 0.5M, 1M, and 4M IDPrior samples improves the average verification accuracy from 94.63\% for CW alone to 95.56\%, 95.81\%, and 96.28\%, respectively. This demonstrates that IDPrior can effectively complement authentic data. At 1M synthetic images, its performance is comparable to VIGFace augmentation, while scaling IDPrior to 4M images yields further gains.}

\vspace{-3mm}
\section{Conclusion}
We presented IDPrior, a framework for synthetic face generation that decouples identity generation from image synthesis. Unlike existing approaches that derive identities through image generation or handcrafted latent distributions, IDPrior learns a diffusion prior over FR embeddings and generates novel identities directly in the embedding space. The generated embeddings can be seamlessly integrated into pretrained identity-conditioned DMs, enabling a modular and scalable generation pipeline.
Extensive experiments demonstrate that IDPrior (DDPM) provides a favorable balance between identity separation, distribution fidelity, and novelty, substantially outperforming PCA- and image-generation-based approaches in this joint evaluation.
When integrated into multiple identity-conditioned DMs, IDPrior generally improves downstream FR performance, while remaining effective for both synthetic-only training and augmentation of authentic data. 

\section*{Acknowledgment}
This research work has been funded by the German Federal Ministry of Education and Research and the Hessen State Ministry for Higher Education, Research and the Arts within their joint support of the National Research Center for Applied Cybersecurity ATHENE and the ARIS Research programmes P2-0250 Metrology and Biometric Systems and P2-0214
Computer Vision.

{
    \small
    \bibliographystyle{ieeenat_fullname}
    \bibliography{main}
}

\newpage

\section*{IJCB Benchmark}
We evaluated on the large-scale IARPA Janus Benchmark–C (IJB-C) \cite{DBLP:conf/icb/MazeADKMO0NACG18}. We used the official 1:1 mixed verification protocol and reported the verification performance as True Acceptance Rates (TAR) at False Acceptance Rates (FAR) of 1e-2, 1e-3, 1e-4 and 1e-5.  The results of IDPrior and other competitors are provided in Table \ref{tab:ijcb}.
\begin{table}[t!]
\centering
\begin{tabular}{lllll}
\hline
            & \multicolumn{4}{l}{IJCB - TPR@FPR} \\ \hline
            & 1e-5    & 1e-4   & 1e-3   & 1e-2   \\ \hline
CW          & 61.41   & 85.32  & 92.71  & 96.73  \\ \hline
DigiFace    & -       & -      & 30.01  & 44.78  \\
IDNet       & -       & -      & 38.85  & 83.25  \\
ExFaceGAN   & -       & -      & 12.92  & 73.28  \\
IDiff-Face  & -       & -      & 23.44  & 69.69  \\
NegFaceDiff & -       & -      & 77.38  & 83.11  \\
DCFace      &         & 77.24  & 87.51  & 94.01  \\
CemiFace    & -       & 83.11  & 90.41  & 95.39  \\
Vec2Face    & -       & 86.10  & 91.04  & 94.92  \\
VIGFace     & -       & 80.00  & 89.22  & 94.99  \\
UIFace      & 81.78   & 88.70  & -      & -      \\
IDPerturb   & 82.28   & 89.49  & 93.71  & 96.56  \\ \hline
IDPrior (ours)    & 82.66   & 90.41  & 94.23  & 96.90  \\ \hline
\end{tabular}
\caption{Verification results on the IJC-B benchmark.}
\label{tab:ijcb}
\end{table}

\section*{Impact of Face recognition backbone and training dataset on IDPrior}
The main paper trains IDPrior on WebFace4M using ElasticFace embeddings for integration with IDiff-Face and IDPerturb, and ArcFace embeddings for integration with Arc2Face and IP-Adapter-FaceID. In this section, we investigate the influence of the face recognition backbone used to extract identity embeddings and the dataset used to train IDPrior.

\begin{figure}[th!]
    \centering
    \begin{subfigure}[b]{\linewidth}
        \centering
        \includegraphics[width=\linewidth]{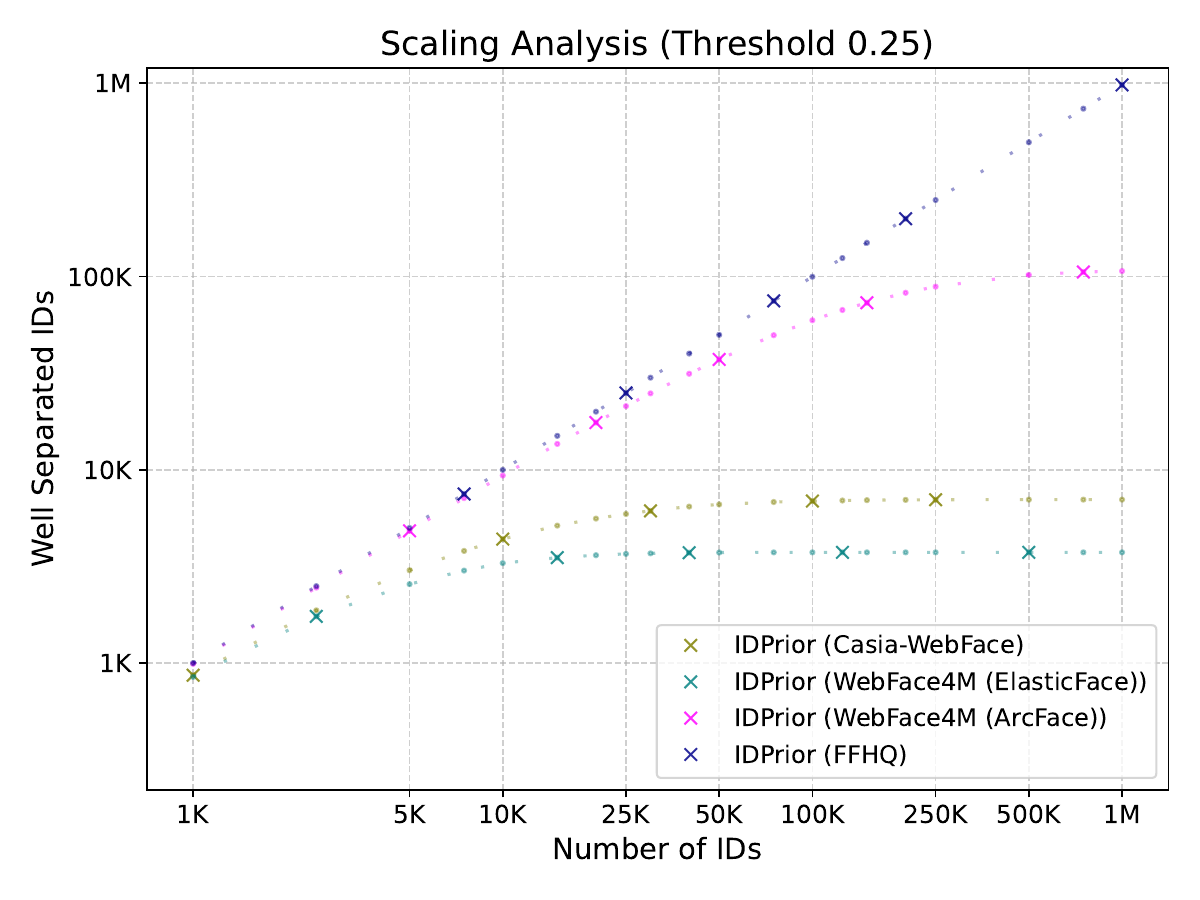}                 
        \caption{}
        \label{fig:enter-label2}
    \end{subfigure}%
    \\
    \begin{subfigure}[b]{\linewidth}
        \centering
        \includegraphics[width=\linewidth]{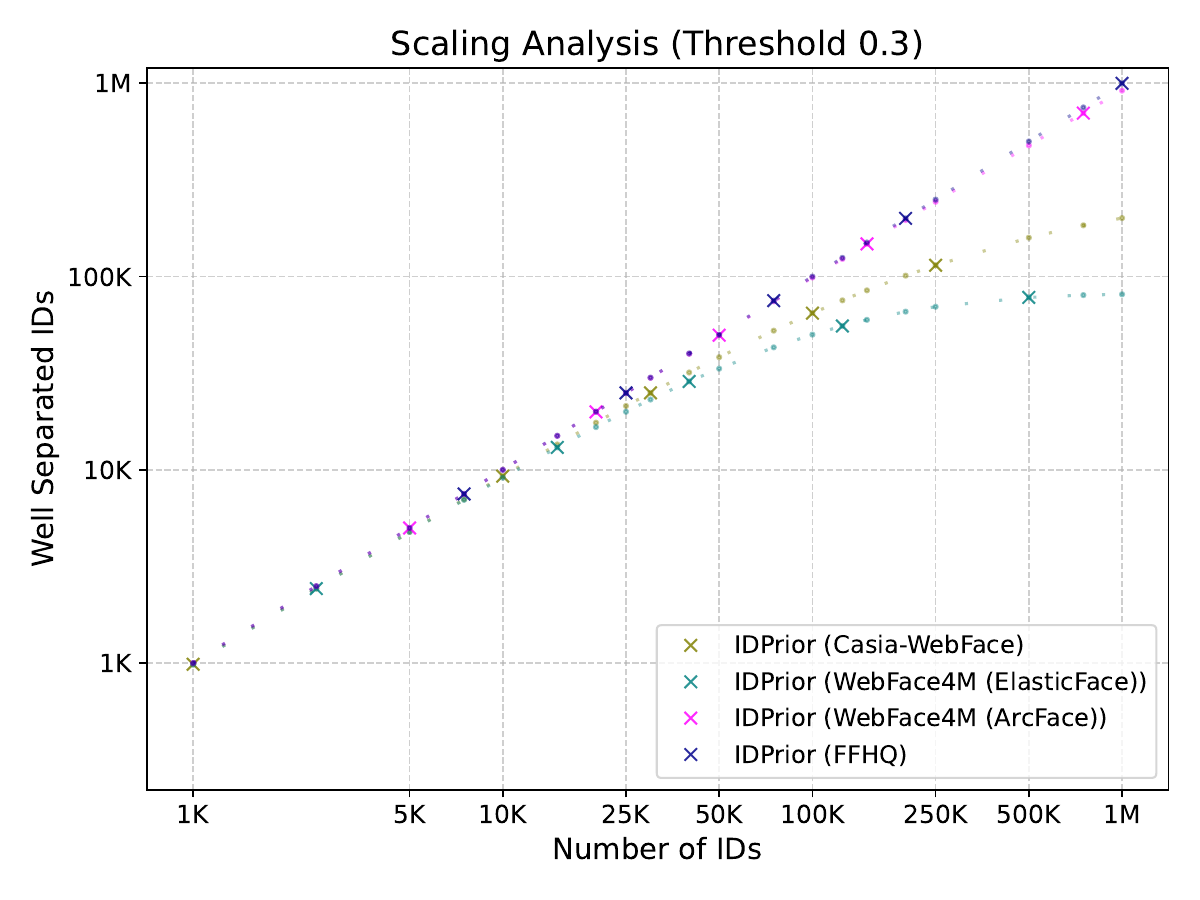}
                             
        \caption{}
    \end{subfigure}
    \caption{
Scalability analysis under different training configurations. The number of well-separated identities vs. the total number of generated identities increases. We compare IDPrior trained on embeddings extracted by ArcFace and ElasticFace, as well as IDPrior trained on embeddings from WebFace4M, CASIA-WebFace, and FFHQ. ArcFace consistently produces more well-separated identities than ElasticFace. Among the training datasets, FFHQ and WebFace4M achieve comparable scalability and substantially outperform CASIA-WebFace..
}
\label{fig:sm_scalability}
\end{figure}

\begin{algorithm}[t]
\caption{Training IDPrior}
\label{alg:training}
\begin{algorithmic}[1]
\Require
Training images $\mathcal{X}=\{x_i\}_{i=1}^{N}$,
pretrained face recognition model $f$,
number of diffusion steps $T$

\State Extract normalized identity embeddings
\[
\mathcal{Z}=\{z_i=f(x_i)\}_{i=1}^{N}
\]

\While{not converged}
    \State Sample a mini-batch $\{z_0\}\subset\mathcal{Z}$
    \State Sample timestep $t\sim\mathcal{U}(\{1,\ldots,T\})$
    \State Sample noise $\epsilon\sim\mathcal{N}(0,I)$
    \State Generate noisy embeddings
    \[
    z_t=\sqrt{\bar{\alpha}_t}z_0+
    \sqrt{1-\bar{\alpha}_t}\epsilon
    \]
    \State Predict noise
    \[
    \hat{\epsilon}=\epsilon_\theta(z_t,t)
    \]
    \State Minimize
    \[
    \mathcal{L}
    =
    \|\epsilon-\hat{\epsilon}\|_2^2
    \]
    \State Update $\theta$
\EndWhile

\State \Return learned identity prior $p_\theta(z)$
\end{algorithmic}
\end{algorithm}

\begin{algorithm}[t]
\caption{Sampling Synthetic Face Dataset with IDPrior}
\label{alg:sampling}
\begin{algorithmic}[1]
\Require
Identity diffusion model $p_\theta(z)$,
pretrained identity-conditioned image generator $p_\phi(x|z)$,
number of identities $N$,
number of images per identity $M$

\State Initialize synthetic dataset $\mathcal{D}\leftarrow\emptyset$

\For{$i=1,\ldots,N$}
    \State Sample a novel identity embedding
    \[
    z_i \sim p_\theta(z)
    \]
    \For{$j=1,\ldots,M$}
        \State Sample a face image conditioned on $z_i$
        \[
        x_{i,j}\sim p_\phi(x|z_i)
        \]
        \State $\mathcal{D}\leftarrow
        \mathcal{D}\cup\{(x_{i,j},i)\}$
    \EndFor
\EndFor

\State \Return $\mathcal{D}$
\end{algorithmic}
\end{algorithm}

\begin{figure}[th!]
    \centering
    \includegraphics[width=1\linewidth]{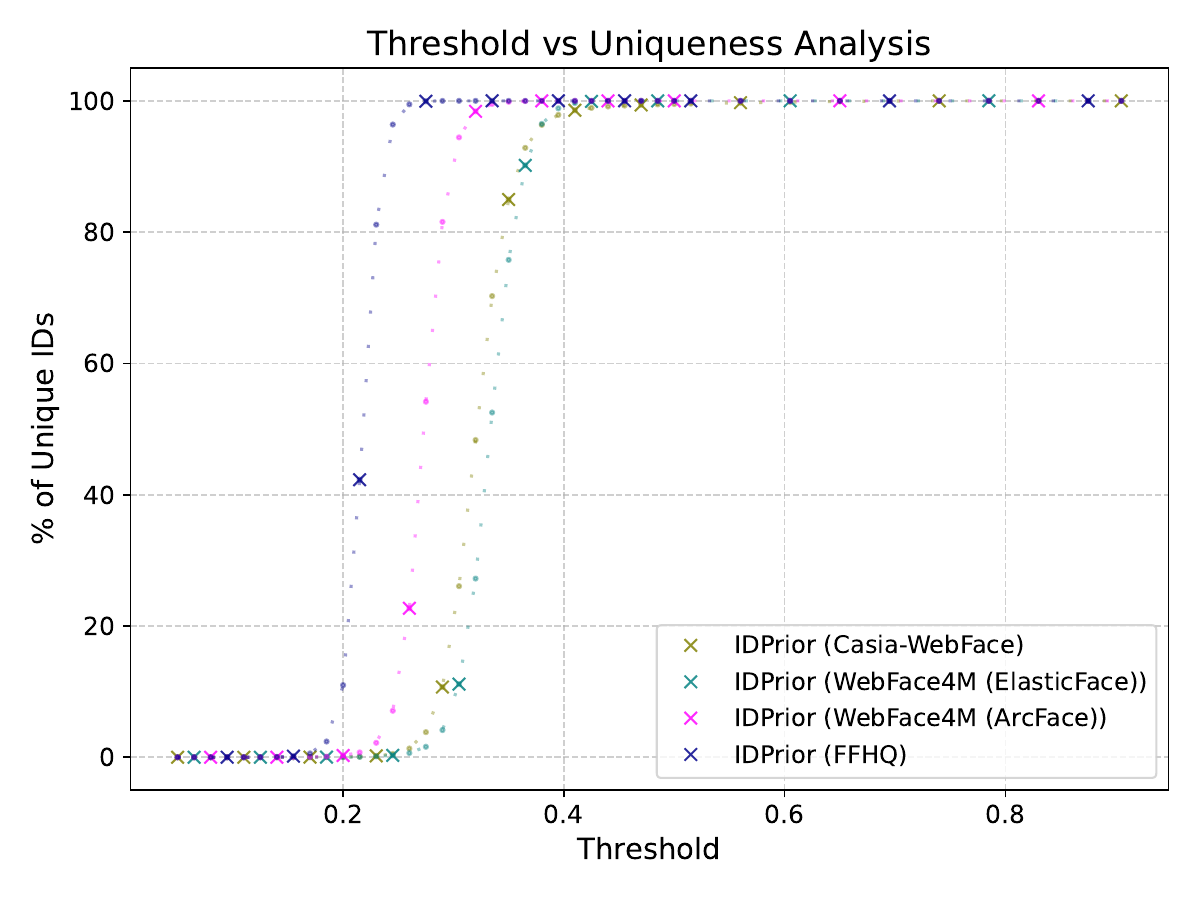}
    \caption{Identity uniqueness under different training configurations. Percentage of unique identities as the cosine-similarity threshold varies from 0 to 1. We compare IDPrior trained using ArcFace and ElasticFace embeddings, as well as IDPrior trained on WebFace4M, CASIA-WebFace, and FFHQ. ArcFace consistently achieves slightly higher identity uniqueness than ElasticFace, while FFHQ and WebFace4M produce comparable results and substantially outperform CASIA-WebFace.}
    \label{fig:sm_uniqness}
\end{figure}

Specifically, we compare IDPrior trained on embeddings extracted by ArcFace~\cite{Deng2020} and ElasticFace~\cite{ElasticFace}. We further evaluate three training datasets with different characteristics: WebFace4M~\cite{DBLP:conf/cvpr/ZhuHDY0CZYLD021}, a large-scale face recognition dataset (4M images of 206K identities), CASIA-WebFace~\cite{DBLP:journals/corr/YiLLL14a}, a relatively smaller face recognition dataset (0.5M images of 10K identities), and FFHQ~\cite{FFHQ}, a high-quality face generation dataset (70K images) without identity labels.

Figure~\ref{fig:sm_uniqness} compares the identity uniqueness obtained by the different training configurations, while Figure~\ref{fig:sm_scalability} reports the scalability analysis.

\subsection*{Effect of the Sampling Strategy}
\begin{figure}[th!]
    \centering
    \includegraphics[width=1\linewidth]{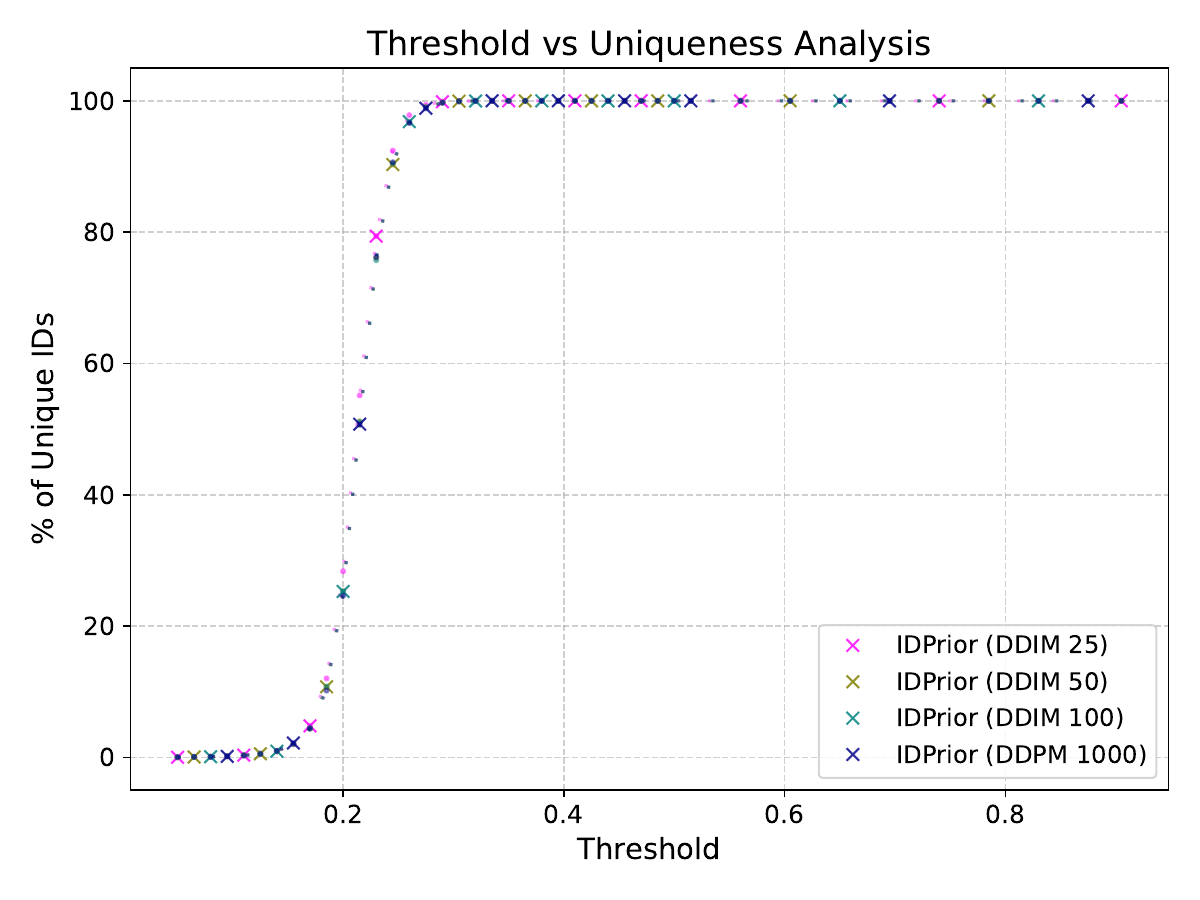}
    \caption{Effect of the sampling strategy on identity uniqueness. Percentage of unique identities as the cosine-similarity threshold varies from 0 to 1 for DDPM (1000 steps) and DDIM with 25, 50, and 100 denoising steps. All sampling strategies produce nearly identical identity uniqueness across the full range of thresholds. All results are reported  using IDPrior trained on embeddings extracted from the ArcFace model of the WebFace4M dataset.}
    \label{fig:sm_uniqness_DDPM}
\end{figure}

In the main paper, IDPrior generates identity embeddings using a DDIM sampler with 50 denoising steps. Here, we compare the standard DDPM sampler with 1000 denoising steps against DDIM using 25, 50, and 100 denoising steps. We evaluate the generated identities using the identity uniqueness (Figure \ref{fig:sm_uniqness_DDPM}) and scalability (Figure \ref{fig:sm_uniqness_DDPM}) metrics introduced in the main paper.

\begin{figure}[t!]
\begin{subfigure}[b]{\linewidth}
    \begin{subfigure}[b]{0.23\linewidth}
        \centering
        \includegraphics[width=\linewidth]{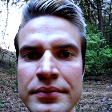}
    \end{subfigure}%
        \begin{subfigure}[b]{0.23\linewidth}
        \centering
        \includegraphics[width=\linewidth]{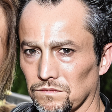}      
    \end{subfigure}%
        \begin{subfigure}[b]{0.23\linewidth}
        \centering
        \includegraphics[width=\linewidth]{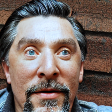}
    \end{subfigure}%
        \begin{subfigure}[b]{0.23\linewidth}
        \centering
        \includegraphics[width=\linewidth]{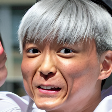}
    \end{subfigure}%
       \caption{Samples of identities generated by Arc2Face~\cite{arc2face} via  PCA approach.
       }
       \label{fig:samples_arc2face}
    \end{subfigure} 
     \begin{subfigure}[b]{\linewidth}
        \begin{subfigure}[b]{0.23\linewidth}
        \centering
        \includegraphics[width=\linewidth]{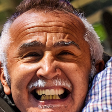}
    \end{subfigure}%
        \begin{subfigure}[b]{0.23\linewidth}
        \centering
        \includegraphics[width=\linewidth]{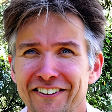}      
    \end{subfigure}%
        \begin{subfigure}[b]{0.23\linewidth}
        \centering
        \includegraphics[width=\linewidth]{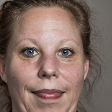}
    \end{subfigure}%
        \begin{subfigure}[b]{0.23\linewidth}
        \centering
        \includegraphics[width=\linewidth]{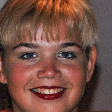}
    \end{subfigure}%
    \caption{Samples of identities generated by Arc2Face~\cite{arc2face} via IDPrior (Ours).} \end{subfigure}
         \begin{subfigure}[b]{\linewidth}
        \begin{subfigure}[b]{0.23\linewidth}
        \centering
        \includegraphics[width=\linewidth]{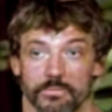}
    \end{subfigure}%
        \begin{subfigure}[b]{0.23\linewidth}
        \centering
        \includegraphics[width=\linewidth]{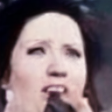}      
    \end{subfigure}%
        \begin{subfigure}[b]{0.23\linewidth}
        \centering
        \includegraphics[width=\linewidth]{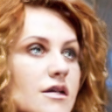}
    \end{subfigure}%
        \begin{subfigure}[b]{0.23\linewidth}
        \centering
        \includegraphics[width=\linewidth]{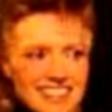}
    \end{subfigure}%
    \caption{Samples of identities generated by IDiff-Face \cite{DBLP:conf/iccv/BoutrosGKD23} via their corresponding approach.} \end{subfigure}
         \begin{subfigure}[b]{\linewidth}
        \begin{subfigure}[b]{0.23\linewidth}
        \centering
        \includegraphics[width=\linewidth]{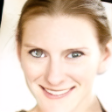}
    \end{subfigure}%
        \begin{subfigure}[b]{0.23\linewidth}
        \centering
        \includegraphics[width=\linewidth]{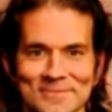}      
    \end{subfigure}%
        \begin{subfigure}[b]{0.23\linewidth}
        \centering
        \includegraphics[width=\linewidth]{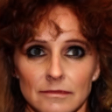}
    \end{subfigure}%
        \begin{subfigure}[b]{0.23\linewidth}
        \centering
        \includegraphics[width=\linewidth]{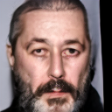}
    \end{subfigure}%
    \caption{Samples of identities generated by IDiff-Face \cite{DBLP:conf/iccv/BoutrosGKD23} via IDPrior (Ours).} 
    \end{subfigure}  
         \begin{subfigure}[b]{\linewidth}
        \begin{subfigure}[b]{0.23\linewidth}
        \centering
        \includegraphics[width=\linewidth]{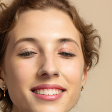}
    \end{subfigure}%
        \begin{subfigure}[b]{0.23\linewidth}
        \centering
        \includegraphics[width=\linewidth]{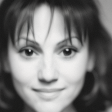}      
    \end{subfigure}%
        \begin{subfigure}[b]{0.23\linewidth}
        \centering
        \includegraphics[width=\linewidth]{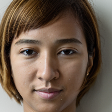}
    \end{subfigure}%
        \begin{subfigure}[b]{0.23\linewidth}
        \centering
        \includegraphics[width=\linewidth]{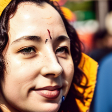}
    \end{subfigure}%
    \caption{Samples of identities generated by IP-Adapter-FaceID \cite{ye2023ip-adapter} via PCA.} 
    \end{subfigure}        
         \begin{subfigure}[b]{\linewidth}
        \begin{subfigure}[b]{0.23\linewidth}
        \centering
        \includegraphics[width=\linewidth]{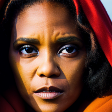}
    \end{subfigure}%
        \begin{subfigure}[b]{0.23\linewidth}
        \centering
        \includegraphics[width=\linewidth]{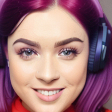}      
    \end{subfigure}%
        \begin{subfigure}[b]{0.23\linewidth}
        \centering
        \includegraphics[width=\linewidth]{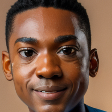}
    \end{subfigure}%
        \begin{subfigure}[b]{0.23\linewidth}
        \centering
        \includegraphics[width=\linewidth]{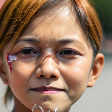}
    \end{subfigure}%
    \caption{Samples of  identities generated by IP-Adapeter-FaceID \cite{ye2023ip-adapter} via IDPrior.} 
    \end{subfigure}
    \vspace{-7mm}
    \caption{Sample images of novel identities generated by different DM models and using different identity generation approaches.  Images are generated from different reference identities.  }
\label{fig:samples}
\vspace{-3mm}
\end{figure}
\begin{figure}[th!]
    \centering
    \begin{subfigure}[b]{\linewidth}
        \centering
        \includegraphics[width=\linewidth]{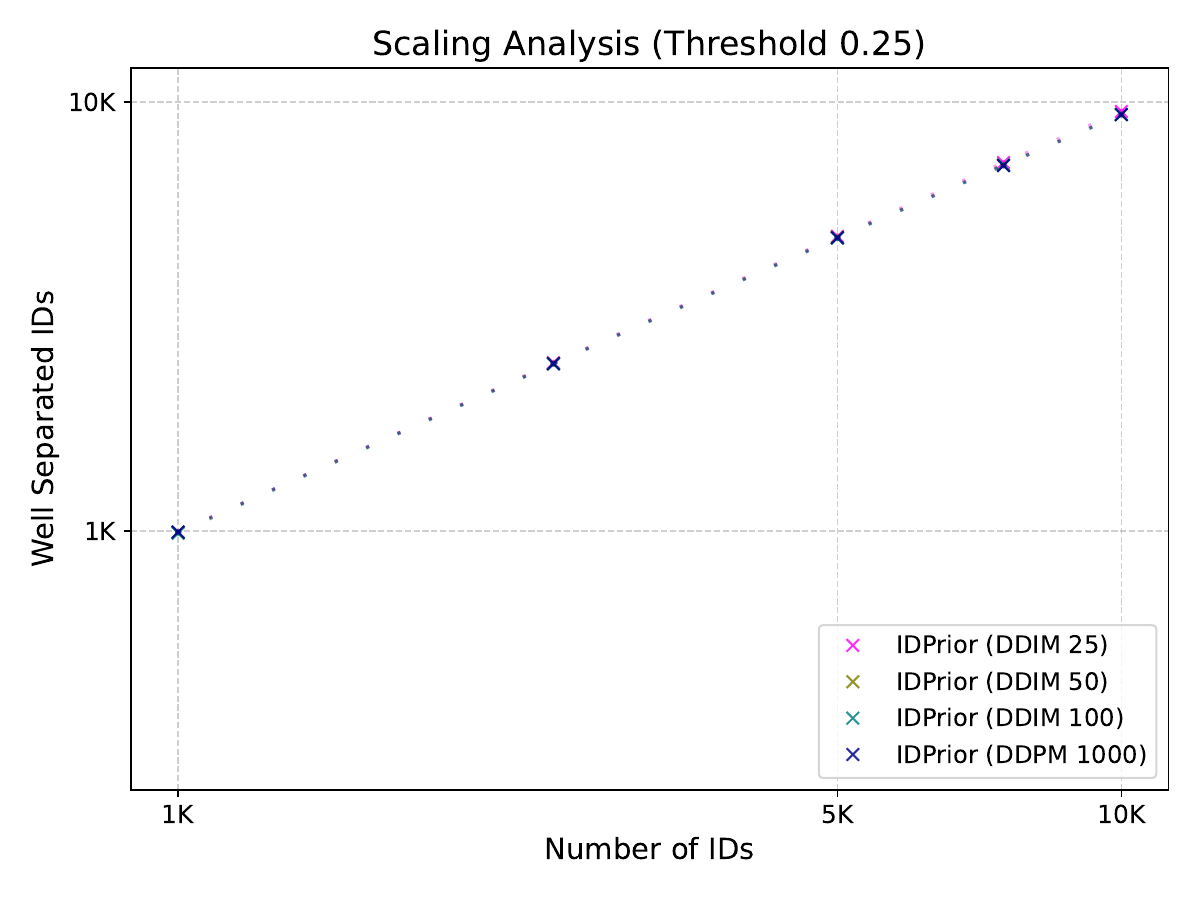}                 
        \caption{}
        \label{fig:enter-label3}
    \end{subfigure}%
    \\
    \begin{subfigure}[b]{\linewidth}
        \centering
        \includegraphics[width=\linewidth]{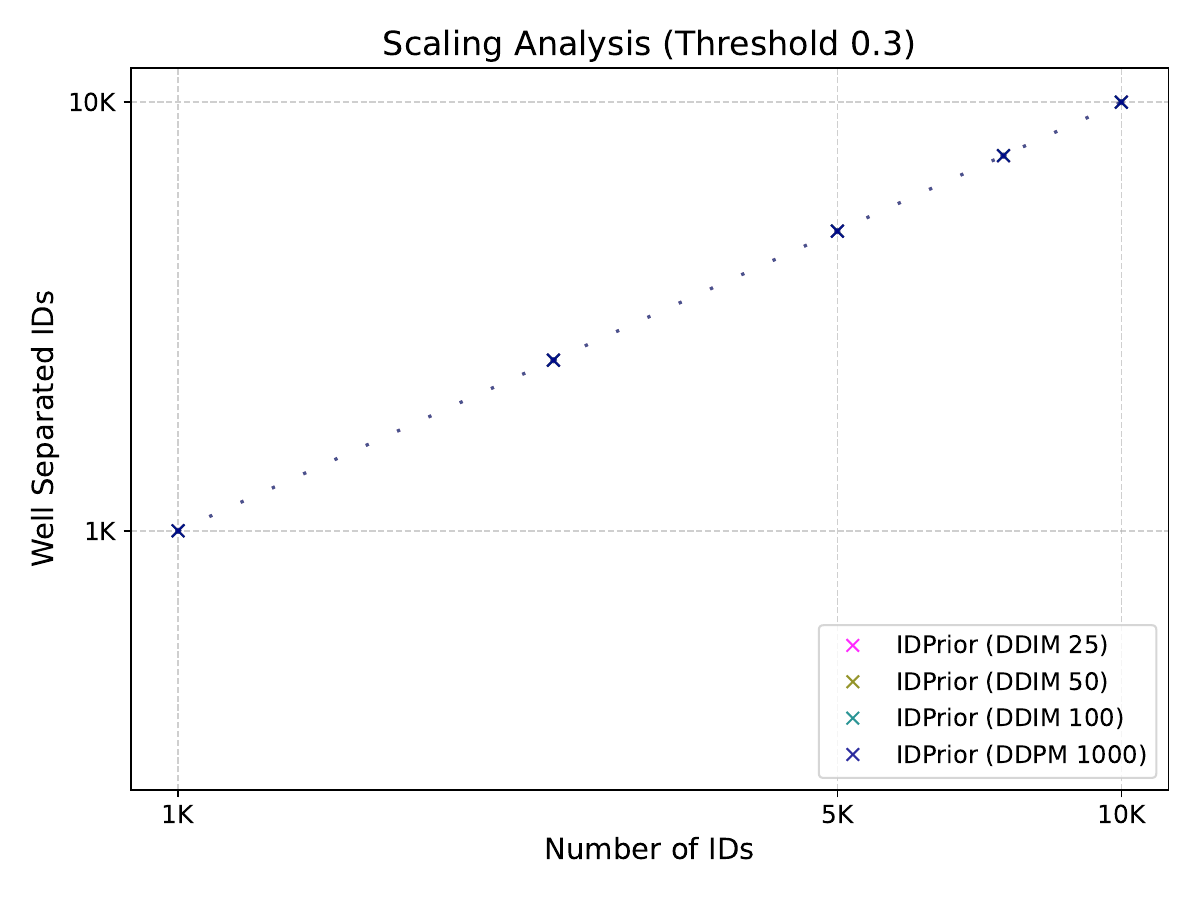}                     
        \caption{}
    \end{subfigure}
    \caption{
Effect of the sampling strategy on scalability. Number of well-separated identities as the number of generated identities increases for DDPM (1000 steps) and DDIM with 25, 50, and 100 denoising steps. DDIM closely matches the scalability of DDPM while requiring substantially fewer denoising steps. All results are reported  using IDPrior trained on embeddings extracted from the ArcFace model of the WebFace4M dataset.
}
\label{fig:sm_scalability_DDPM}
\end{figure}

\subsection*{Unconditional Identity Generation with IP-Adapter-FaceID}

\begin{figure}[th!]
    \centering
    \begin{subfigure}[b]{\linewidth}
        \centering
        \includegraphics[width=\linewidth]{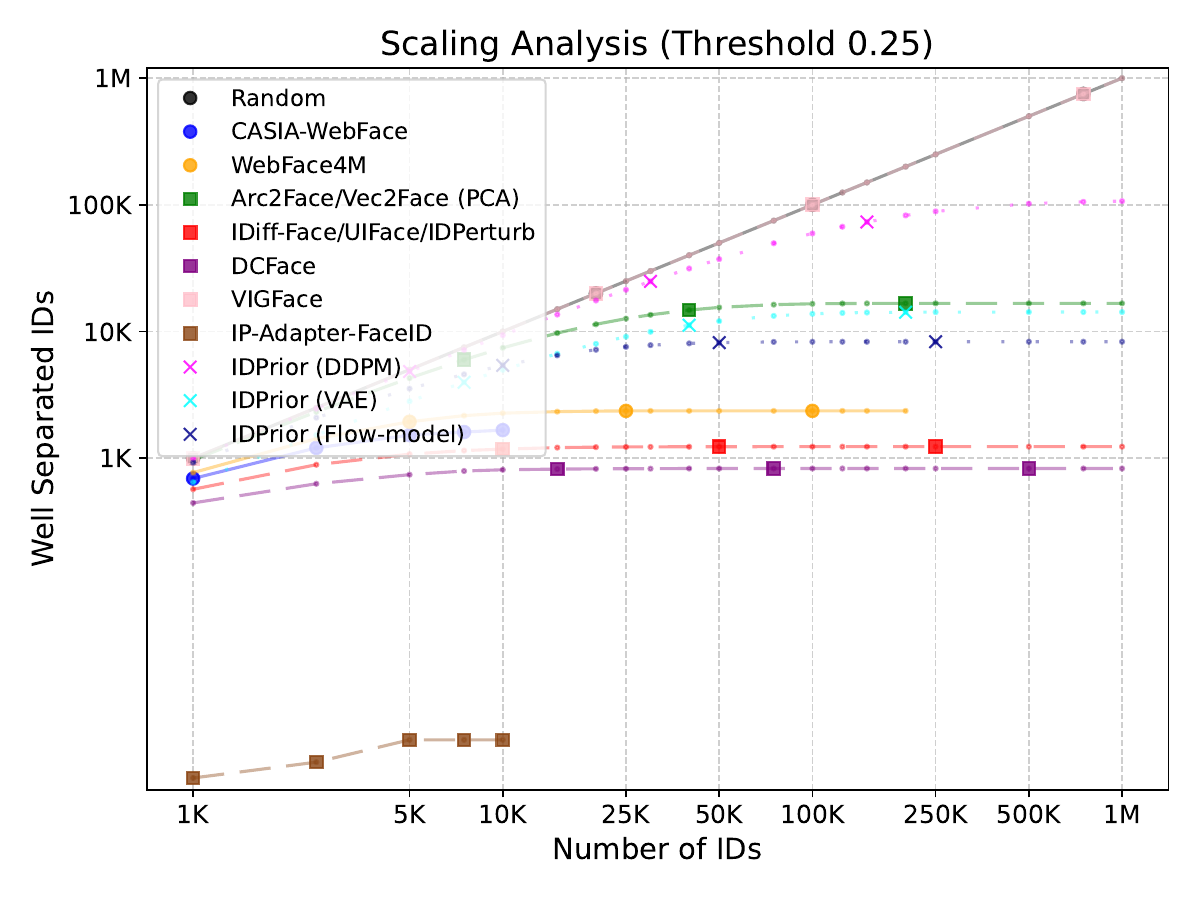}                 
        \caption{}
        \label{fig:enter-label}
    \end{subfigure}%
    \\
    \begin{subfigure}[b]{\linewidth}
        \centering
        \includegraphics[width=\linewidth]{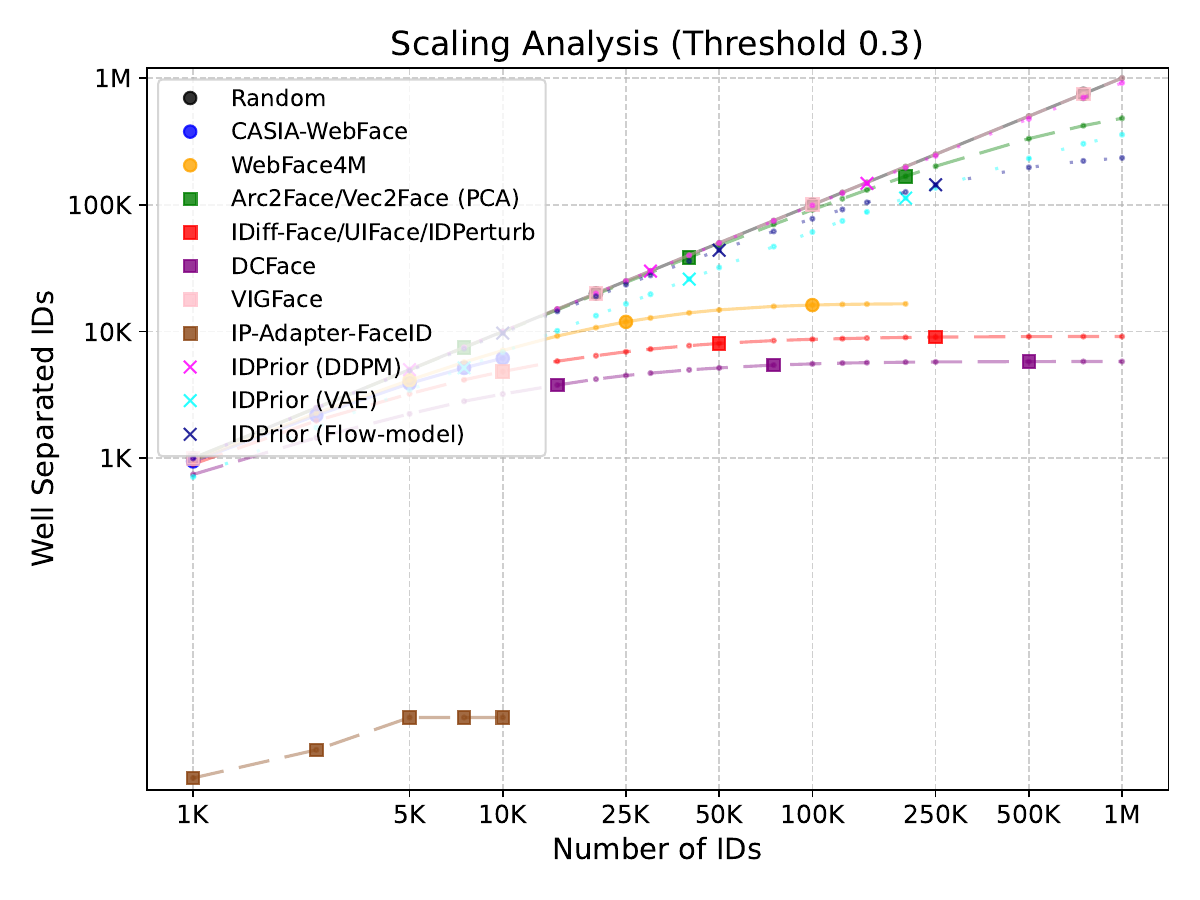}                 
                             
        \caption{}
    \end{subfigure}
    \caption{
Scalability evaluation, including unconditional IP-Adapter-FaceID. The number of well-separated identities vs. the total number of generated identities increases. Unconditional sampling from IP-Adapter-FaceID quickly saturates, producing fewer than 3\% well-separated identities due to severe identity collisions. These results justify the use of a PCA-based baseline for IP-Adapter-FaceID in the main paper.
}
\label{fig:sm_scalability_ip}
\end{figure}

\begin{figure}[th!]
    \centering
    \includegraphics[width=1\linewidth]{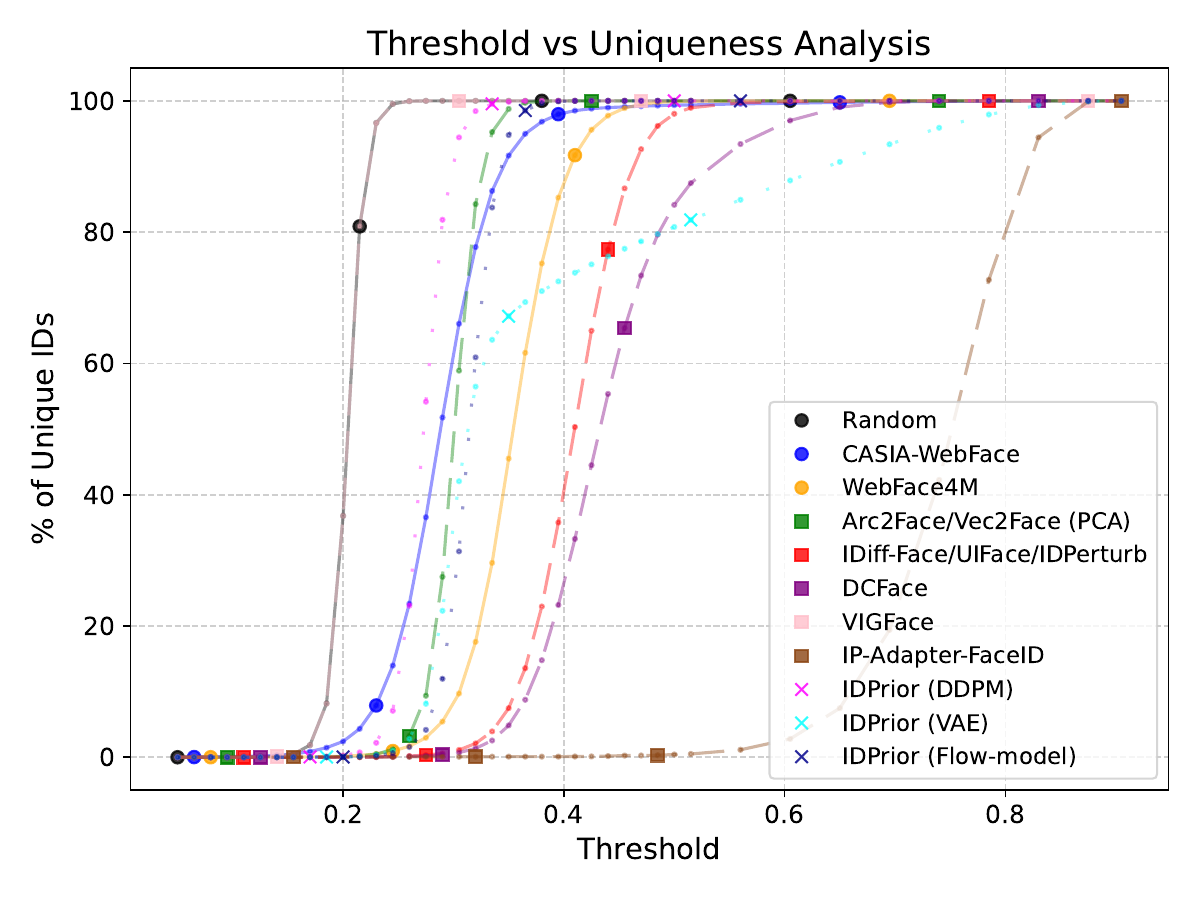}
    \caption{Identity uniqueness including unconditional IP-Adapter-FaceID. Percentage of unique identities as the cosine-similarity threshold varies from 0 to 1. Unconditional sampling from IP-Adapter-FaceID produces fewer than 3\% unique identities across most thresholds, substantially underperforming other approaches.}
    \label{fig:sm_uniqness_ip}
\end{figure}

In the main paper, we compare IDPrior with a PCA-based identity generation baseline for IP-Adapter-FaceID instead of unconditional image generation. Here, we justify this design choice by extending the identity uniqueness and scalability analyses to include identities obtained by unconditional sampling from IP-Adapter-FaceID. Each generated image is treated as a distinct identity and evaluated using the identity uniqueness and scalability metrics, Figure \ref{fig:sm_uniqness_ip}, Figure \ref{fig:sm_scalability_ip} and Tables \ref{tab:Id_uniqueness_details_sm} and \ref{tab:Id_uniqueness_details_sm2}.

\section*{Social Impact}
\label{sec}

This work is part of an ongoing research effort to enable the development of face recognition systems using fully synthetic data, thereby reducing the reliance on authentic biometric data and mitigating the associated legal, ethical, and privacy concerns. By decoupling identity generation from image synthesis, the proposed framework facilitates the scalable generation of synthetic training data while remaining compatible with existing identity-conditioned image generation models.

We envision this research supporting the development of face recognition systems for legitimate applications, such as secure authentication, identity verification, and access control, within applicable legal and regulatory frameworks. At the same time, we recognize that face recognition technologies may be susceptible to misuse. We therefore advocate their responsible development and deployment, accompanied by appropriate safeguards, transparency, and compliance with relevant laws and ethical standards. The proposed method is intended solely to advance privacy-preserving and ethically responsible biometric research and should not be used for unlawful, malicious, or discriminatory purposes.

\section*{Use of existing assets and face recognition evaluation benchmarks}
\label{sec:app_g}

\subsection*{Use of Existing Assets}

The results of state-of-the-art face recognition models reported in the main paper are taken directly from their respective publications, unless otherwise stated. We utilize the publicly available pretrained IDiff-Face~\cite{DBLP:conf/iccv/BoutrosGKD23} model trained on CASIA-WebFace, which is released under the Attribution-NonCommercial-ShareAlike 4.0 International (CC BY-NC-SA 4.0) license and available at \url{https://github.com/EduardaCaldeira/NegFaceDiff}.

Arc2Face~\cite{arc2face} is available under the MIT license at \url{https://github.com/foivospar/Arc2Face}. We use the official pretrained image generation model and train the PCA-based identity generator ourselves, as neither a pretrained PCA model nor synthetic identities are publicly available. The PCA training and sampling procedure follows the protocol described in the original Arc2Face paper.

The pretrained IP-Adapter-FaceID~\cite{ye2023ip-adapter} model is obtained from the official repository (\url{https://github.com/tencent-ailab/IP-Adapter}). We use the default text prompt \texttt{"best quality, high quality"} and the default negative prompt \texttt{"monochrome, lowres, bad anatomy, worst quality, low quality"} for image generation.

We provide the following details for all datasets used throughout this work.
 
\textbf{CASIA-WebFace \cite{DBLP:conf/iclr/LinHXMZD25,DBLP:conf/cvpr/Kim0L22}:}
CASIA-WebFace consists of 494,141 face images from 10,757 different identities. A preprocessed (aligned and cropped) version of CASIA-WebFace is publicly available in the InsightFace (\url{https://insightface.ai/}) repository  under Dataset-Zoo
\url{https://github.com/deepinsight/insightface/tree/master/recognition/_datasets_ }. 
The code and the databases of InsightFace are under MIT licence (\url{https://github.com/deepinsight/insightface/blob/master/LICENSE}).

\textbf{WebFace4M} \cite{DBLP:conf/cvpr/ZhuHDY0CZYLD021} is subset of WebFace260M contining 4M images with 206K identities. The dataset was officially obtained from \cite{DBLP:conf/cvpr/ZhuHDY0CZYLD021}. 

\textbf{FFHQ \cite{FFHQ}}
Flickr-Faces-HQ Dataset (FFHQ) consists of 70,000 high-quality PNG images. The data is available for research under \url{https://github.com/NVlabs/ffhq-dataset}.

\subsection*{Evaluation benchmarks}

This section presents the description and license information of the FR evaluation datasets used in our work.

\textbf{AgeDB-30 \cite{AgeDB30Database}:} AgeDB is an in-the-wild dataset for age-invariant face verification evaluation, containing 16,488 images of 568 identities. Every image is annotated with respect to the identity, age, and gender attribute. In our case, we report the performance for AgeDB-30 (30 years age gap) as it is the most reported and challenging subset of AgeDB. More details on the collection process can be found in \cite{AgeDB30Database} and the details on the license are presented in \url{https://ibug.doc.ic.ac.uk/resources/agedb/}.

\textbf{LFW \cite{LFWDatabase}}: Labeled Faces in the Wild (LFW) is an unconstrained face verification dataset. The LFW contains 13,233 images of 5749 identities collected from the web. The LFW is licensed under CC-BY-4.0, and more information on database creation can be found in \cite{LFWDatabase} and \url{http://vis-www.cs.umass.edu/lfw/}.

\textbf{CFP-FP \cite{CFPFPDatabase}:} Celebrities in Frontal-Profile in the Wild (CFP-FP) \cite{CFPFPDatabase} dataset addresses the comparison between frontal and profile faces. CFP-FP dataset contains 7,000 images across 500 identities, where 10 frontal and 4 profile image per identity.
More information can be found in \cite{CFPFPDatabase} and \url{http://www.cfpw.io/}.

\textbf{CALFW \cite{CALFWDatabase}:} The Cross-age LFW (CALFW) dataset \cite{CALFWDatabase} is based on LFW with a focus on comparison pairs with the age gap, however not as large as AgeDB-30. Age gap distribution of the CALFW is provided in \cite{CALFWDatabase}. It contains 3000 genuine comparisons, and the negative pairs are selected of the same gender and race to reduce the effect of attributes. The detailed information on database creation can be found in \cite{CALFWDatabase} and \url{http://whdeng.cn/CALFW/}.

\textbf{CPLFW \cite{CPLFWDatabase}:} The Cross-Pose LFW (CPLFW) dataset \cite{CPLFWDatabase} is based on LFW with a focus on comparison pairs with pose differences. CPLFW contains 3000 genuine comparisons, while the negative pairs are selected of the same gender and race. More information can be found in \cite{CPLFWDatabase} and \url{http://whdeng.cn/CPLFW/}.

\begin{table*}[]
\centering
\caption{
Scalability of identity generation from 1K to 1M identities, which extends the results of the main submission. We report the percentage of unique identities at thresholds 0.25, 0.30, and 0.40. IDPrior consistently maintains the highest uniqueness across all scales, substantially outperforming image-generation-based and PCA-based approaches. We additionally reported results of authentic datasets, CASIA-WebFace (10K identities) and WebFace4M (206K  identities). Identities of unconditional IP-Adapter-FaceID sampling saturated at 10K identities. Given the page size, the other models' results are provided in Table \ref{tab:Id_uniqueness_details_sm2}.
}
\resizebox{0.70\linewidth}{!}{%
\begin{tabular}{llllllllll}
method                                    & samples & r@0.25   & r@0.30   & r@0.40   & method                                        & samples & 0.25  & 0.30  & 0.40  \\ \hline
\multirow{20}{*}{Arc2Face/Vec2Face (PCA)} & 1000    & 97.10  & 99.90  & 100.00 & \multirow{20}{*}{DCFace}                      & 1000    & 44.40 & 74.50 & 94.30 \\ 
                                          & 2500    & 92.96  & 99.80  & 100.00 &                                               & 2500    & 25.16 & 58.08 & 90.84 \\
                                          & 5000    & 85.78  & 99.62  & 100.00 &                                               & 5000    & 14.84 & 44.84 & 87.16 \\
                                          & 7500    & 79.83  & 99.39  & 100.00 &                                               & 7500    & 10.57 & 37.56 & 85.41 \\
                                          & 10000   & 74.35  & 99.15  & 100.00 &                                               & 10000   & 8.11  & 32.10 & 83.75 \\
                                          & 15000   & 64.83  & 98.85  & 100.00 &                                               & 15000   & 5.47  & 25.19 & 81.09 \\
                                          & 20000   & 57.03  & 98.28  & 100.00 &                                               & 20000   & 4.13  & 21.04 & 78.92 \\
                                          & 25000   & 50.52  & 97.77  & 100.00 &                                               & 25000   & 3.31  & 17.99 & 77.26 \\
                                          & 30000   & 45.15  & 97.14  & 100.00 &                                               & 30000   & 2.76  & 15.64 & 75.52 \\
                                          & 40000   & 36.92  & 96.19  & 100.00 &                                               & 40000   & 2.08  & 12.46 & 72.59 \\
                                          & 50000   & 31.00  & 95.33  & 100.00 &                                               & 50000   & 1.66  & 10.32 & 70.08 \\
                                          & 75000   & 21.72  & 93.25  & 100.00 &                                               & 75000   & 1.11  & 7.25  & 65.29 \\
                                          & 100000  & 16.55  & 91.25  & 100.00 &                                               & 100000  & 0.83  & 5.55  & 61.43 \\
                                          & 125000  & 13.30  & 89.28  & 100.00 &                                               & 125000  & 0.66  & 4.51  & 58.15 \\
                                          & 150000  & 11.10  & 87.42  & 100.00 &                                               & 150000  & 0.55  & 3.78  & 55.40 \\
                                          & 200000  & 8.34   & 83.86  & 100.00 &                                               & 200000  & 0.42  & 2.86  & 50.88 \\
                                          & 250000  & 6.67   & 80.55  & 100.00 &                                               & 250000  & 0.33  & 2.30  & 47.35 \\
                                          & 500000  & 3.34   & 66.48  & 100.00 &                                               & 500000  & 0.17  & 1.16  & 36.48 \\
                                          & 750000  & 2.22   & 56.10  & 100.00 &                                               & 750000  & 0.11  & 0.77  & 30.38 \\
                                          & 1000000 & 1.67   & 48.15  & 100.00 &                                               & 1000000 & 0.08  & 0.58  & 26.34 \\ \hline
\multirow{20}{*}{IDPrior (Ours)}          & 1000    & 99.10  & 100.00 & 100.00 & \multirow{20}{*}{IDiff-Face/UIFace/IDPerturb} & 1000    & 56.90 & 89.90 & 99.70 \\
                                          & 2500    & 98.28  & 99.96  & 100.00 &                                               & 2500    & 35.56 & 79.04 & 99.40 \\
                                          & 5000    & 96.70  & 99.98  & 100.00 &                                               & 5000    & 21.54 & 64.52 & 98.80 \\
                                          & 7500    & 95.16  & 99.96  & 100.00 &                                               & 7500    & 15.33 & 55.44 & 98.15 \\
                                          & 10000   & 93.44  & 99.91  & 100.00 &                                               & 10000   & 11.79 & 48.35 & 97.60 \\
                                          & 15000   & 90.75  & 99.87  & 100.00 &                                               & 15000   & 8.09  & 38.86 & 96.55 \\
                                          & 20000   & 87.84  & 99.78  & 100.00 &                                               & 20000   & 6.12  & 32.22 & 95.28 \\
                                          & 25000   & 85.41  & 99.75  & 100.00 &                                               & 25000   & 4.91  & 27.68 & 94.25 \\
                                          & 30000   & 82.88  & 99.69  & 100.00 &                                               & 30000   & 4.10  & 24.21 & 93.25 \\
                                          & 40000   & 78.53  & 99.56  & 100.00 &                                               & 40000   & 3.08  & 19.34 & 91.57 \\
                                          & 50000   & 74.48  & 99.43  & 100.00 &                                               & 50000   & 2.47  & 16.11 & 89.79 \\
                                          & 75000   & 66.28  & 99.17  & 100.00 &                                               & 75000   & 1.65  & 11.31 & 85.87 \\
                                          & 100000  & 59.39  & 98.90  & 100.00 &                                               & 100000  & 1.23  & 8.68  & 82.43 \\
                                          & 125000  & 53.71  & 98.67  & 100.00 &                                               & 125000  & 0.99  & 7.03  & 79.40 \\
                                          & 150000  & 48.81  & 98.44  & 100.00 &                                               & 150000  & 0.82  & 5.92  & 76.75 \\
                                          & 200000  & 41.21  & 97.97  & 100.00 &                                               & 200000  & 0.62  & 4.49  & 71.90 \\
                                          & 250000  & 35.47  & 97.51  & 100.00 &                                               & 250000  & 0.49  & 3.61  & 67.95 \\
                                          & 500000  & 20.40  & 95.40  & 100.00 &                                               & 500000  & 0.25  & 1.82  & 54.37 \\
                                          & 750000  & 14.07  & 93.47  & 100.00 &                                               & 750000  & 0.16  & 1.22  & 46.14 \\
                                          & 1000000 & 10.68  & 91.64  & 100.00 &                                               & 1000000 & 0.12  & 0.91  & 40.43 \\ \hline
\multirow{20}{*}{Random}                  & 1000    & 100.00 & 100.00 & 100.00 & \multirow{20}{*}{IP-Adapter-FaceID}           & 1000    & 0.30  & 0.30  & 0.40  \\
                                          & 2500    & 100.00 & 100.00 & 100.00 &                                               & 2500    & 0.16  & 0.20  & 0.28  \\
                                          & 5000    & 100.00 & 100.00 & 100.00 &                                               & 5000    & 0.12  & 0.18  & 0.22  \\
                                          & 7500    & 100.00 & 100.00 & 100.00 &                                               & 7500    & 0.08  & 0.12  & 0.15  \\
                                          & 10000   & 100.00 & 100.00 & 100.00 &                                               & 10000   & 0.06  & 0.09  & 0.11  \\
                                          & 15000   & 100.00 & 100.00 & 100.00 &                                               & -       & -     & -     & -     \\
                                          & 20000   & 100.00 & 100.00 & 100.00 &                                               & -       & -     & -     & -     \\
                                          & 25000   & 100.00 & 100.00 & 100.00 &                                               & -       & -     & -     & -     \\
                                          & 30000   & 99.99  & 100.00 & 100.00 &                                               & -       & -     & -     & -     \\
                                          & 40000   & 99.99  & 100.00 & 100.00 &                                               & -       & -     & -     & -     \\
                                          & 50000   & 99.98  & 100.00 & 100.00 &                                               & -       & -     & -     & -     \\
                                          & 75000   & 99.97  & 100.00 & 100.00 &                                               & -       & -     & -     & -     \\
                                          & 100000  & 99.96  & 100.00 & 100.00 &                                               & -       & -     & -     & -     \\
                                          & 125000  & 99.96  & 100.00 & 100.00 &                                               & -       & -     & -     & -     \\
                                          & 150000  & 99.96  & 100.00 & 100.00 &                                               & -       & -     & -     & -     \\
                                          & 200000  & 99.95  & 100.00 & 100.00 &                                               & -       & -     & -     & -     \\
                                          & 250000  & 99.94  & 100.00 & 100.00 &                                               & -       & -     & -     & -     \\
                                          & 500000  & 99.88  & 100.00 & 100.00 &                                               & -       & -     & -     & -     \\
                                          & 750000  & 99.82  & 100.00 & 100.00 &                                               & -       & -     & -     & -     \\
                                          & 1000000 & 99.76  & 100.00 & 100.00 &                                               & -       & -     & -     & -     \\ \hline
\multirow{16}{*}{WebFace4M}               & 1000    & 77.00  & 95.80  & 99.90  & \multirow{16}{*}{CASIA-WebFace}               & 1000    & 69.20 & 94.40 & 99.70 \\
                                          & 2500    & 57.12  & 90.48  & 99.84  &                                               & 2500    & 48.28 & 87.24 & 99.40 \\
                                          & 5000    & 38.82  & 82.56  & 99.60  &                                               & 5000    & 30.46 & 77.82 & 98.80 \\
                                          & 7500    & 28.91  & 76.35  & 99.39  &                                               & 7500    & 21.47 & 68.87 & 98.51 \\
                                          & 10000   & 22.68  & 70.83  & 99.32  &                                               & 10000   & 16.67 & 61.50 & 98.17 \\
                                          & 15000   & 15.55  & 61.43  & 98.99  &                                               & -       & -     & -     & -     \\
                                          & 20000   & 11.77  & 53.68  & 98.64  &                                               & -       & -     & -     & -     \\
                                          & 25000   & 9.46   & 47.66  & 98.29  &                                               & -       & -     & -     & -     \\
                                          & 30000   & 7.88   & 42.62  & 97.94  &                                               & -       & -     & -     & -     \\
                                          & 40000   & 5.91   & 35.10  & 97.23  &                                               & -       & -     & -     & -     \\
                                          & 50000   & 4.73   & 29.62  & 96.54  &                                               & -       & -     & -     & -     \\
                                          & 75000   & 3.15   & 21.05  & 94.98  &                                               & -       & -     & -     & -     \\
                                          & 100000  & 2.37   & 16.17  & 93.54  &                                               & -       & -     & -     & -     \\
                                          & 125000  & 1.89   & 13.09  & 92.06  &                                               & -       & -     & -     & -     \\
                                          & 150000  & 1.58   & 10.97  & 90.66  &                                               & -       & -     & -     & -     \\
                                          & 200000  & 1.18   & 8.26   & 88.08  &                                               &         & -     & -     & -    \\ \hline

\end{tabular}}

\label{tab:Id_uniqueness_details_sm}
\end{table*}

\begin{table}[]
\centering
\caption{
Scalability of identity generation from 1K to 1M identities, which extends the results of the main submission. This table is complementary to Table \ref{tab:Id_uniqueness_details_sm}.
}
\resizebox{0.90\linewidth}{!}{%
\begin{tabular}{lllll}
\hline
method                          & Samples & r@0.25 & r@0.30 & r@0.40 \\ \hline
\multirow{20}{*}{IDPrior (Flow-model)} & 1000    & 92.30  & 99.60  & 100.00 \\
                                      & 2500    & 83.56  & 98.88  & 100.00 \\
                                      & 5000    & 70.84  & 98.10  & 99.98  \\
                                      & 7500    & 61.25  & 97.68  & 99.99  \\
                                      & 10000   & 54.05  & 97.04  & 99.99  \\
                                      & 15000   & 43.20  & 96.04  & 99.99  \\
                                      & 20000   & 35.91  & 94.82  & 100.00 \\
                                      & 25000   & 30.28  & 93.64  & 99.99  \\
                                      & 30000   & 26.00  & 92.43  & 99.99  \\
                                      & 40000   & 20.15  & 89.80  & 99.99  \\
                                      & 50000   & 16.34  & 87.54  & 99.99  \\
                                      & 75000   & 11.04  & 82.25  & 99.99  \\
                                      & 100000  & 8.30   & 77.61  & 99.99  \\
                                      & 125000  & 6.64   & 73.46  & 99.99  \\
                                      & 150000  & 5.54   & 69.62  & 99.99  \\
                                      & 200000  & 4.15   & 63.04  & 99.99  \\
                                      & 250000  & 3.32   & 57.52  & 99.98  \\
                                      & 500000  & 1.66   & 39.46  & 99.97  \\
                                      & 750000  & 1.11   & 29.57  & 99.96  \\
                                      & 1000000 & 0.83   & 23.44  & 99.95  \\ \hline

\multirow{20}{*}{IDPrior (VAE)} & 1000    & 64.70 & 71.20 & 79.00  \\
                                & 2500    & 61.00 & 69.84 & 77.96  \\
                                & 5000    & 56.30 & 69.06 & 77.44  \\
                                & 7500    & 52.99 & 68.93 & 77.29  \\
                                & 10000   & 49.87 & 68.58 & 77.15  \\
                                & 15000   & 44.54 & 67.52 & 76.57  \\
                                & 20000   & 39.99 & 66.67 & 76.25  \\
                                & 25000   & 36.41 & 66.10 & 76.03  \\
                                & 30000   & 33.16 & 65.59 & 75.92  \\
                                & 40000   & 28.03 & 64.79 & 75.74  \\
                                & 50000   & 24.15 & 63.99 & 75.59  \\
                                & 75000   & 17.72 & 62.42 & 75.29  \\
                                & 100000  & 13.80 & 60.97 & 75.00  \\
                                & 125000  & 11.21 & 59.66 & 74.79  \\
                                & 150000  & 9.41  & 58.56 & 74.67  \\
                                & 200000  & 7.10  & 56.44 & 74.42  \\
                                & 250000  & 5.69  & 54.51 & 74.24  \\
                                & 500000  & 2.85  & 46.38 & 73.56  \\
                                & 750000  & 1.90  & 40.41 & 73.19  \\
                                & 1000000 & 1.42  & 35.72 & 72.95  \\ \hline

\multirow{20}{*}{VIGFace}       & 1000    & 100.00 & 100.00 & 100.00 \\
                                & 2500    & 100.00 & 100.00 & 100.00 \\
                                & 5000    & 100.00 & 100.00 & 100.00 \\
                                & 7500    & 100.00 & 100.00 & 100.00 \\
                                & 10000   & 100.00 & 100.00 & 100.00 \\
                                & 15000   & 100.00 & 100.00 & 100.00 \\
                                & 20000   & 100.00 & 100.00 & 100.00 \\
                                & 25000   & 100.00 & 100.00 & 100.00 \\
                                & 30000   & 100.00 & 100.00 & 100.00 \\
                                & 40000   & 100.00 & 100.00 & 100.00 \\
                                & 50000   & 99.99  & 100.00 & 100.00 \\
                                & 75000   & 99.99  & 100.00 & 100.00 \\
                                & 100000  & 99.98  & 100.00 & 100.00 \\
                                & 125000  & 99.98  & 100.00 & 100.00 \\
                                & 150000  & 99.97  & 100.00 & 100.00 \\
                                & 200000  & 99.96  & 100.00 & 100.00 \\
                                & 250000  & 99.95  & 100.00 & 100.00 \\
                                & 500000  & 99.88  & 100.00 & 100.00 \\
                                & 750000  & 99.82  & 100.00 & 100.00 \\
                                & 1000000 & 99.77  & 100.00 & 100.00 \\ \hline
\end{tabular}}
\label{tab:Id_uniqueness_details_sm2}
\end{table}

\begin{figure}
    \centering
    \includegraphics[width=1\linewidth]{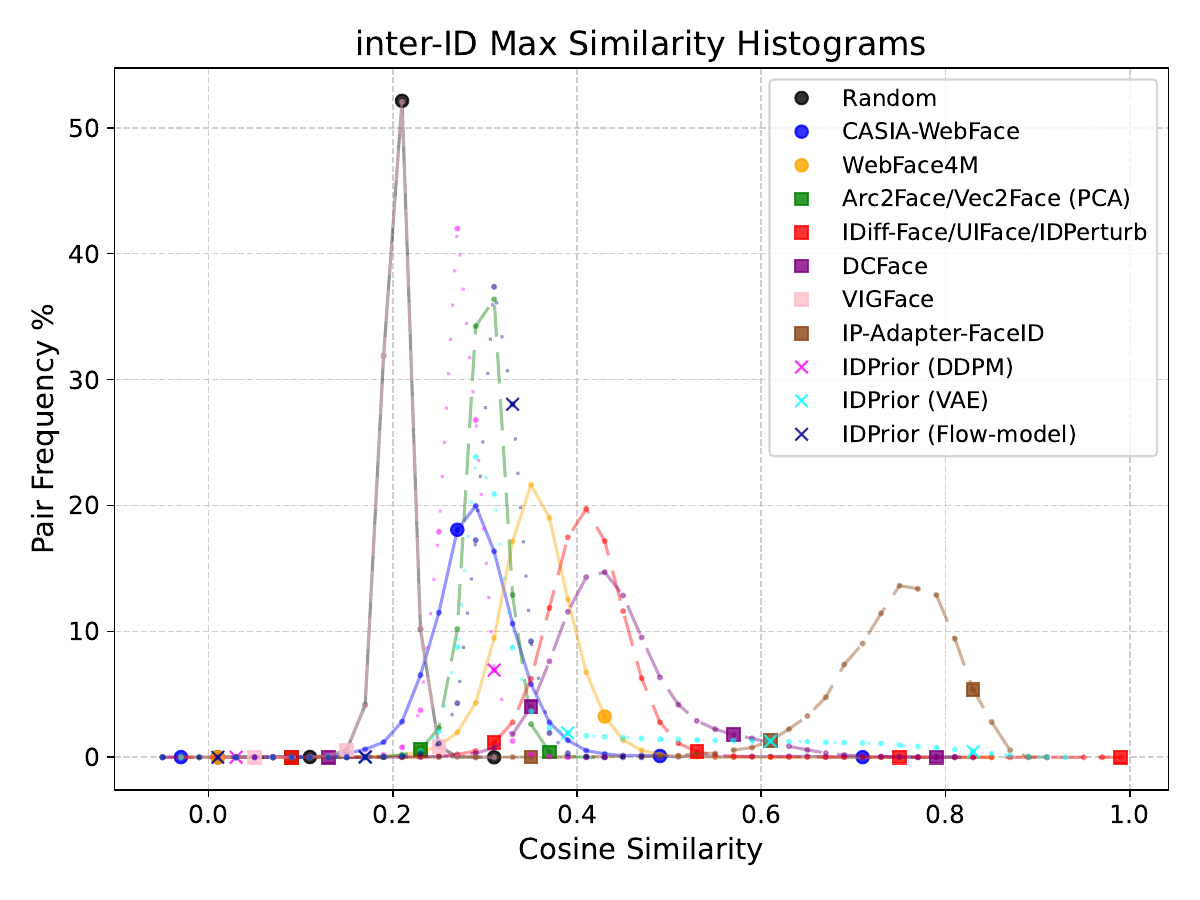}
    \caption{Histogram of the maximum cosine similarity between each generated identity embedding and all other generated identities within each dataset. Lower similarity indicates better inter-class separation and fewer identity collisions. IDPrior shifts the distribution toward lower similarities compared with image-generation-based and PCA-based approaches, demonstrating its ability to generate more distinct identities.}
    \label{fig:enter-label1}
\end{figure}

\end{document}